\documentclass[sigconf, nonacm]{acmart}
\setcopyright{none}
\renewcommand\footnotetextcopyrightpermission[1]{}

\usepackage{booktabs}
\usepackage{multirow}
\usepackage{xcolor}
\usepackage{algorithm}
\usepackage{algpseudocode}
\usepackage{amsmath}
\usepackage{pifont}
\usepackage{graphicx}

\usepackage{xspace}
\newcommand{\ourmethod}{\textsc{Attic}\xspace}
\newcommand{\ourmethodfull}{\textsc{Attic-KV}\xspace}

\usepackage{colortbl}
\definecolor{oursbg}{HTML}{FFF7E3}
\definecolor{headbg}{HTML}{E3EAF2}
\definecolor{gainc}{HTML}{D9481C}
\definecolor{demph}{gray}{0.5}
\newcommand{\gain}[1]{\textsubscript{\textcolor{gainc}{$\uparrow$#1}}}
\newcommand{\loss}[1]{\textsubscript{\textcolor{demph}{$\downarrow$#1}}}
\newlength{\gainw}
\newcommand{\gslot}[1]{\makebox[\gainw][l]{#1}}
\newcommand{\oursrow}{\rowcolor{oursbg}}
\newcommand{\headrow}{\rowcolor{headbg}}
\newlength{\gw}
\newcommand{\www}[2]{#2}
\newcommand{\lbcatpos}{t}
\newif\ifgeneral\generaltrue

\begin{document}

\title{Rehearse Everything, Remember Nothing: Attic-KV Rehearses What Will Be Read}

\author{Zhiyun Shi}
\affiliation{\institution{Nanyang Technological University}\country{Singapore}}
\email{ZHIYUN004@e.ntu.edu.sg}

\begin{abstract}
Many key--value (KV) caches are compressed before anyone knows what will be asked of them: a document cached for retrieval, a prompt prefix shared across requests, the memory of a long conversation.
The prevailing approach scores KV entries by \textit{rehearsal}: the model rereads the context and keeps the entries it attends to, assuming that the more completely a cache rehearses its context, the better it remembers it.
We show that under tight budgets this assumption backfires: \textbf{rehearse everything, remember nothing}.
At a 3\% keep ratio, rereading the whole context keeps 31.5 of 96.5 points on RULER, and on LongBench's natural-text tasks it falls below methods that rehearse nothing at all. %
The cause is that \textbf{a cache keeps what it rehearses}: rereading spreads the budget across the whole context, so the answer's own entries survive at little more than chance. %
Like a student before an exam, a cache remembers more by testing itself than by rereading.
Two principles follow: \textbf{rehearse what will be read}, and \textbf{rehearse as much as there is}.
We instantiate them as \textbf{\ourmethodfull} (\ourmethod for short), a training-free rehearsal in which the model quizzes itself with question--answer pairs that quote the context, alongside anchor tokens in a content-adaptive amount.
Changing only the rehearsal lifts three hosts that score it in three different ways: \ourmethod alone is the best training-free method in all eight settings we test on RULER and LongBench's natural-text tasks, and plugged into the gradient-based KVgrad and the trained RestoreKV+, it raises them by up to 17.1 and 28.1 points. %
Its advantage grows as the budget shrinks, reaching 41.9 points over full rereading at a 3\% keep ratio, and on long documents it compresses faster than rereading them. %
\end{abstract}


\begin{teaserfigure}
  \centering
  \includegraphics[width=\textwidth]{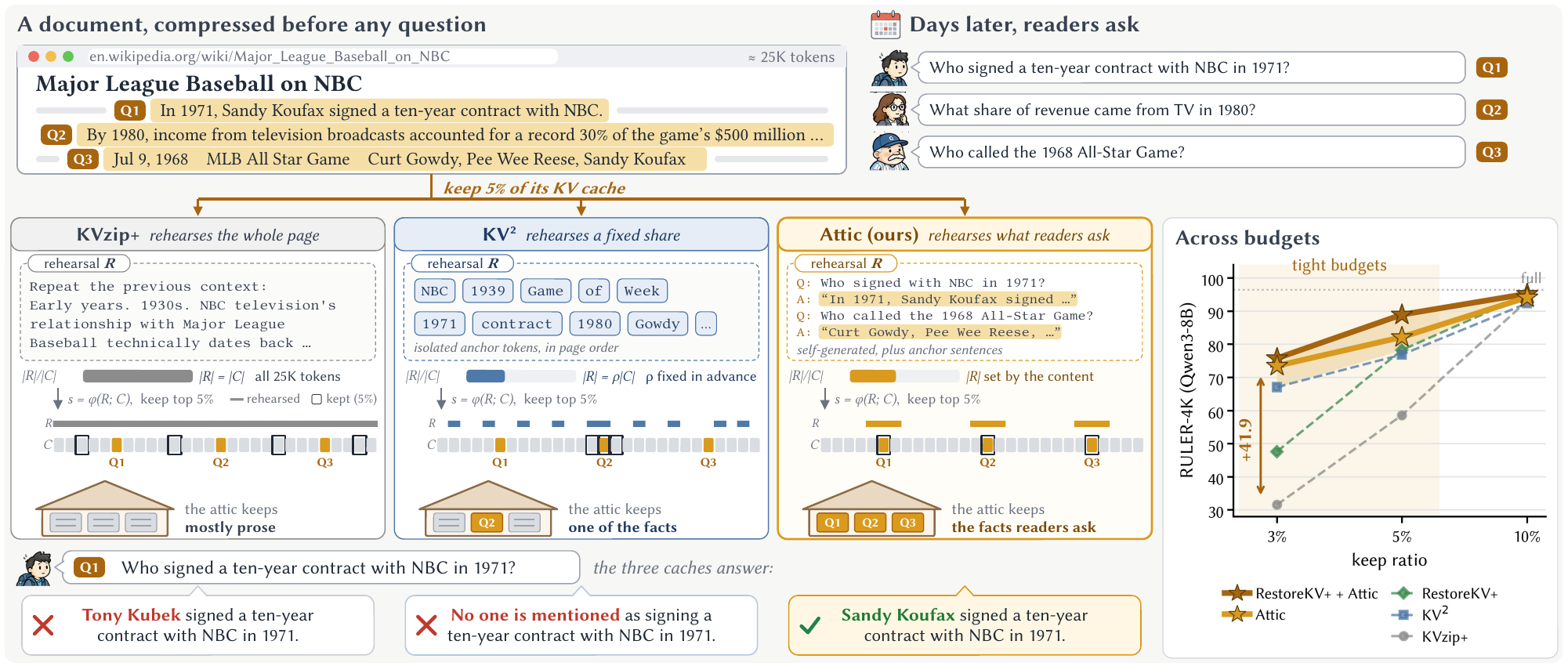}
  \vspace{-4mm}
  \caption{\textbf{Rehearse everything, remember nothing.} A document, here a Wikipedia page, is compressed once, to 5\% of its KV cache, before any question arrives.
Each method feeds a rehearsal $R$ after the page and keeps the 5\% of entries that $R$ attends to most, so a compressed cache, like a small attic, keeps what it rehearses.
KVzip+ rehearses the whole page and keeps mostly prose; KV$^2$ rehearses a fixed share of isolated anchors and keeps only some facts; \ourmethod rehearses self-generated question--answer pairs and anchor sentences, as much as the page holds, and keeps the facts readers later ask about.
Answers are verbatim outputs of Qwen3-8B on this LooGLE page. %
(\textbf{\textit{Right}}) RULER-4K score versus keep ratio on Qwen3-8B: the gap opens at tight budgets and reaches 41.9 points over KVzip+ at 3\%.} %
\label{fig:teaser}

\end{teaserfigure}

\maketitle

\section{Introduction}\label{sec:intro}

When Watson learns that Sherlock Holmes does not know that the Earth travels round the Sun, Holmes explains that a man's brain is like a little empty attic~\cite{holmes1887}.
A fool stocks it with every sort of lumber he comes across, until the knowledge that might be useful is crowded out; a skilful workman takes in nothing but the tools that may help him in his work, and of those he keeps a large assortment.
For, as Holmes warns, ``it is a mistake to think that that little room has elastic walls.''
A compressed key--value (KV) cache is such an attic: it holds a small, fixed fraction of a context's keys and values, and what it holds decides which questions the model can still answer.

Many KV caches are filled before the questions are known.
Retrieval-augmented and search systems precompute the caches of documents and reuse them across requests~\cite{promptcache,ragcache,turborag,cacheblend}; serving systems share the cache of a long prompt prefix across users; assistants keep the cache of a long conversation for turns that have not happened yet.
Each such cache is compressed once and then serves many queries that are unknown when it is compressed~\cite{arckv,cartridgespp}.
Compression must therefore be \textit{query-agnostic} and \textit{one-shot}, and because memory grows with the number of caches, it pays off only when it is extreme: a 5\% keep ratio fits 20 times as many caches in the same memory, a 30\% ratio only about 3 times (Section~\ref{sec:setting}). %

The prevailing paradigm for this setting is \textbf{rehearsal scoring}, introduced by KVzip~\cite{kvzip}.
Before any question arrives, the model \textit{rehearses} the context, rereading it while attending to the cache; each KV entry is scored by the attention it receives, and the top-scoring entries are kept.
Its successors refine the scorer (KVzip+ and KVzap~\cite{kvzap}, Fast KVzip~\cite{fastkvzip}, EchoPress~\cite{echopress}, KVgrad~\cite{kvgrad}) or what a retained subset can express (RestoreKV~\cite{restorekv}), yet all of them rehearse the same target: the entire context.
KVzip argues explicitly for full-context rehearsal~\cite{kvzip}, and the whole line shares its default assumption: the more completely a cache rehearses its context, the more faithfully it preserves it.
Concurrent work, KV$^2$~\cite{kvsquared}, rehearses only a fixed fraction of the context, its anchor tokens; but like every rehearsal before it, what it rehearses is a copy of the context.

\begin{figure}[t]
\centering
\includegraphics[width=\columnwidth]{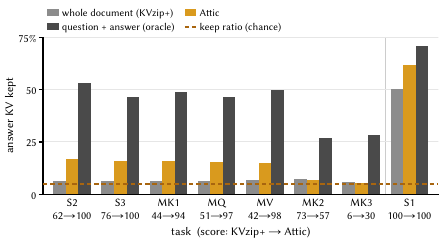}
\caption{\textbf{Rehearse everything, remember nothing.} Share of each answer's KV entries, over all layers and heads, that the cache keeps (Qwen3-8B, RULER-4K development split, the eight needle tasks, keep ratio 5\%). Whole-document rehearsal keeps the answer no more often than any other token (dashed line); \ourmethod keeps it on the tasks it answers, and an oracle that rehearses the gold answer shows how much a 5\% cache can hold. Below each task: its score with whole-document rehearsal $\rightarrow$ with \ourmethod.} %
\label{fig:probe}
\end{figure}

We show that this assumption fails exactly where compression matters: \textbf{rehearse everything, remember nothing}.
On RULER-4K~\cite{ruler}, Qwen3-8B scores 96.5 without compression, but full-context rehearsal (KVzip+) at a 3\% keep ratio keeps only 31.5; on the natural-text tasks of LongBench it falls below methods that rehearse nothing at all (28.6 vs.\ up to 36.5). %
Where does the budget go? For each question, we measure how much of its answer's KV survives compression: at a 5\% keep ratio, full-context rehearsal keeps 5.7--7.1\% of it on seven of eight needle tasks, no more than of any other token (Figure~\ref{fig:probe}). %
The cause is that \textbf{a cache keeps what it rehearses} (Section~\ref{sec:rehearsal}): rehearsed positions cover about 5\% of the context yet draw a fifth to a third of the retained budget, and when the model rehearses only a list of facts, its score on each task tracks the coverage of that list. %
Capacity is not the bottleneck: a 5\% cache scored by rehearsing the answer loses almost nothing (93.5 vs.\ 94.7 uncompressed). %
A full-context rehearsal spends the budget on fluent prose that no query reads and crowds out the few facts that every query reads, and the tighter the budget, the harder the crowding.

The failure has a familiar shape.
A student preparing for an exam can reread the chapter, reread only what they highlighted, or test themselves on it.
A large review of study techniques rates the first two of low utility and the third of high utility~\cite{dunlosky2013improving}: practice that produces answers beats seeing the material again~\cite{roediger2006testing}.
Existing rehearsals are the first two strategies: full-context rehearsal rereads the whole context, and KV$^2$ rereads its highlights.
Neither rehearses what a reader will read out of the context, and a compressed cache, like a student, remembers what it practises.
What counts as practice is specific: rehearsing a question without its answer keeps a cache that scores 47.0, against 93.5 when the answer is rehearsed with it, because attention lands on the answer only while the answer is produced (Section~\ref{sec:p1}). %
Query-agnostic compression is thus first a question of how a cache should study: on Qwen3-8B RULER-4K, replacing the KVzip+ scorer with KVgrad's gradient attribution adds 13.4 points at a 5\% keep ratio and 22.0 at 3\%, while replacing its rehearsal with ours adds 23.5 and 41.9. %

Two principles follow.
\textbf{Rehearse what will be read}: a cache should rehearse what future queries will read out, not the whole context; rehearsing only the facts that can be asked already solves the fact-sparse tasks that full-context rehearsal fails (Section~\ref{sec:p1}).
\textbf{Rehearse as much as there is}: the amount of rehearsal should follow how much readable content a context holds, not a fraction fixed in advance; at an equal average amount, a content-adaptive amount beats a fixed fraction (78.0 vs.\ 74.6; Section~\ref{sec:p2}). %
Under these principles, prior methods become special cases rather than rivals: both fix the amount before reading the context, KVzip at the whole of it and KV$^2$ at a fraction $\rho$, and a content-adaptive rehearsal approaches KVzip as the budget loosens.

Guided by these principles, we propose \ourmethodfull (\ourmethod for short), named after Holmes's brain-attic.
\ourmethod builds the rehearsal from three parts: \textbf{self-generated QA rehearsal}, in which the model writes the questions it expects to be asked and answers them by quoting the context; \textbf{anchors}~\cite{keydiff}, which decide \textit{what} else to rehearse; and the length of the sentences that contain them, which decides \textit{how much} (\textbf{content-adaptive amount}).
\ourmethod is training-free and changes only the rehearsal, so it plugs into any rehearsal-scoring \textit{host method} without altering its scoring (``+ \ourmethod''), and it accepts the released restore tokens of RestoreKV without retraining.
Changing only the rehearsal lifts three hosts that score it in three different ways: by attention, by gradient attribution, and with trained restore tokens.
Alone, \ourmethod is the best training-free method in all eight settings we test (RULER, and the 14 natural-text tasks of LongBench), and at a 3\% keep ratio on RULER-4K it scores 73.4, 6.4 points above KV$^2$ and 41.9 above KVzip+, a lead that grows as the budget shrinks (Figure~\ref{fig:teaser}, right). %
Plugged into KVgrad, it raises the host by up to 17.1 points, and KVgrad + \ourmethod beats every existing training-free method in all eight settings; plugged into RestoreKV+, it raises the host by up to 28.1 points and beats every existing method in all six settings where restore tokens exist (Table~\ref{tab:main}). %
Because it rehearses only part of each context, its compression time grows more slowly with length than full rehearsal's: it matches KVzip+ on a 32K-token context and takes 27\% less on a 64K-token one. %
We make three contributions:
\begin{itemize}
\item \textbf{\textit{Paradigm Reformulation}:} We identify full-context rehearsal, the shared default of rehearsal scoring, as the cause of its failure under tight budgets; we show that \textit{a cache keeps what it rehearses}, so that what a cache remembers depends more on what it rehearses than on how it scores the rehearsal; we observe that every existing rehearsal, the whole context or its anchors, rereads the context instead of practising what will be read out of it; and we recast query-agnostic compression as deciding what and how much to rehearse under two principles, \textit{rehearse what will be read} and \textit{rehearse as much as there is}, with KVzip and KV$^2$ as fixed-amount special cases.
\item \textbf{\textit{Practical Solution}:} We propose \ourmethod, a training-free rehearsal built from self-generated QA, anchors and a content-adaptive amount, which serves alone or as a plug-in to existing hosts and composes with released restore tokens.
\item \textbf{\textit{Experimental Evaluation}:} On RULER with four LLMs and the natural-text tasks of LongBench with two, \ourmethod is the best training-free method, every host it enters beats every existing method that trains as little as it does, and the margin widens as the budget tightens; on LooGLE, where each document is compressed once and answers all of its questions, \ourmethod with restore tokens leads on both short- and long-dependency questions (Section~\ref{sec:experiments}).
\end{itemize}

\section{Problem Setting}\label{sec:setting}

\paragraph{Compress once, answer many.}
Many KV caches are built before anyone knows what will be asked of them. A retrieval-augmented or search system prefills each document of its corpus once and stores the cache for every later request that retrieves it~\cite{promptcache,ragcache,turborag,cacheblend}; a serving system shares the cache of a long system prompt or few-shot prefix across users; an assistant keeps the cache of a long conversation or a tool output for turns that have not happened yet. Two properties are common to all of them.
\textbf{(I) One compression, many unknown queries.} A cache is compressed once and then serves every future query on its context, none of which is known when the cache is compressed.
\textbf{(II) Memory grows with the number of caches.} At keep ratio $r$, a fixed memory holds $1/r$ times as many caches as it holds full ones: 20 times at $r=5\%$, but only about 3 times at $r=30\%$. %
Compression pays off at a few percent, and 3--5\% is the regime we target. %

\paragraph{Query-agnostic, one-shot compression.}
Let $C=(c_1,\dots,c_n)$ be a document of $n$ tokens.
Prefilling $C$ yields, at every layer $l\in[L]$ and KV head $h\in[H]$, a key--value pair $(\mathbf{k}^{l,h}_i,\mathbf{v}^{l,h}_i)$ for each position $i\in[n]$.
We call each pair an \textit{entry} $e=(l,h,i)$ and write $\mathcal{E}$ for the set of all $LHn$ entries.
A compressor keeps a subset $\mathcal{S}\subseteq\mathcal{E}$ of size $r|\mathcal{E}|$, ranked jointly across layers and heads, and the model $f$ then answers every query from the retained cache $\mathrm{KV}_{\mathcal{S}}$.
The goal is
\begin{equation}\label{eq:objective}
\max_{\mathcal{S}\subseteq\mathcal{E},\;|\mathcal{S}|=r|\mathcal{E}|}\;
\mathbb{E}_{(q,a)\sim\mathcal{Q}_C}\Big[\,U\big(f(q\mid \mathrm{KV}_{\mathcal{S}}),\,a\big)\Big],
\end{equation}
where $\mathcal{Q}_C$ is the distribution of future questions $q$ about $C$ with answers $a$, and $U$ is the task metric.
The compression is \textit{query-agnostic}: $\mathcal{Q}_C$ is unknown, so $\mathcal{S}$ must be computed from $C$ alone.
It is \textit{one-shot}: a single $\mathcal{S}$ serves every draw from $\mathcal{Q}_C$.

\begin{figure}[t]
\centering
\includegraphics[width=\columnwidth]{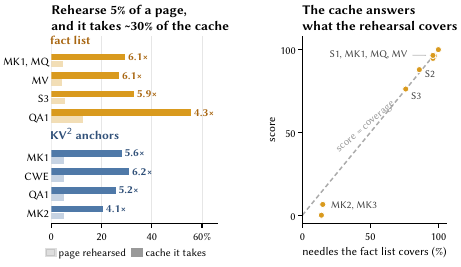}
\caption{\textbf{A cache keeps what it rehearses} (Qwen3-8B, RULER-4K development split, keep ratio 5\%). (\textbf{\textit{Left}}) Share of the document that a rehearsal covers (light) and share of the kept KV entries, over all layers and heads, that fall on those positions (dark), for a semantic rehearsal (fact list) and a geometric one (KV$^2$ anchors). (\textbf{\textit{Right}}) Rehearsing only the fact list: needle coverage of the list against the score. MK/MQ/MV: multi-key/-query/-value needles; S: single needle; CWE: common words; QA: question answering.} %
\label{fig:enrich}
\end{figure}

\begin{figure*}[t]
  \centering
  \includegraphics[width=\textwidth]{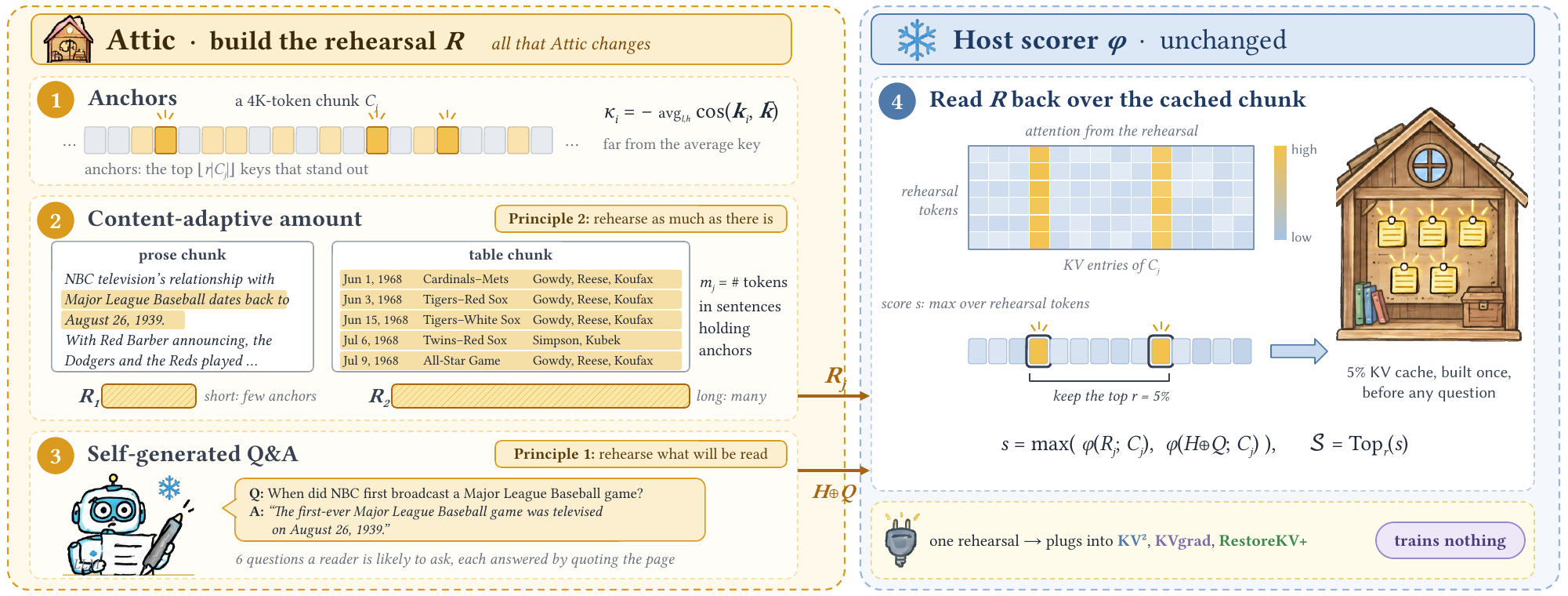}
  \caption{\textbf{Overview of \ourmethod.} (\textbf{\textit{Left}}) \ourmethod builds the rehearsal $R$, the only part of compression it changes: \ding{182} the keys that stand out from the chunk's average key are the anchors; \ding{183} the anchors are expanded to $m_j$ tokens, the length of the sentences that hold them, so a prose chunk is rehearsed briefly and a table of facts at length (Principle~2); \ding{184} the model writes the questions a reader is likely to ask and answers them by quoting the page (Principle~1). (\textbf{\textit{Right}}) The host's scorer $\phi$ reads the rehearsal back over the cached chunk, scores every KV entry by the attention it receives (heat map illustrative), and keeps the top $r$. Text is from the LooGLE page of Figure~\ref{fig:teaser}; the Q\&A pair is the first one the model wrote for it. The same rehearsal plugs into KV$^2$, KVgrad and RestoreKV+, and nothing is trained.} %
  \label{fig:framework}
\end{figure*}

\paragraph{The rehearsal-scoring paradigm.}
The prevailing answer to Eq.~\eqref{eq:objective}, introduced by KVzip~\cite{kvzip}, replaces the unknown queries with a \textbf{rehearsal}.
A rehearsal text $R=(u_1,\dots,u_m)$ is prefilled after $C$ on the full cache, and each entry is scored by the strongest attention it receives from $R$:
\begin{equation}\label{eq:score}
s_R(l,h,i)\;=\;\max_{t\in[m]}\;\max_{g\in\mathcal{G}(h)}\;\tilde{a}^{\,l,g}_{t,i},
\end{equation}
where $\mathcal{G}(h)$ is the set of query heads that share KV head $h$, and $\tilde{a}^{\,l,g}_{t,i}$ is the attention weight from rehearsal token $u_t$ to context position $i$ after the KVzip+ normalization~\cite{kvzap}, which rescales it by the size of the entry's output contribution $\|W_O\mathbf{v}^{l,h}_i\|$ relative to the hidden state of $u_t$.
The kept set is $\mathcal{S}=\mathrm{Top}_r(s_R)$, the $r|\mathcal{E}|$ highest-scoring entries.
Several rehearsals combine entry-wise: $s_{R_1\oplus R_2}=\max(s_{R_1},s_{R_2})$.

Within this paradigm, a method is defined by its rehearsal, and a rehearsal by two choices: its \textit{content} and its \textit{amount} $|R|/n$.
KVzip and its successors rehearse the whole document, $R=C$, behind a short repeat instruction and chunk by chunk for long documents~\cite{kvzip,kvzap,kvgrad,restorekv}; their amount is always 1.
KV$^2$~\cite{kvsquared} rehearses \textbf{anchors}, $R=C[\mathcal{A}_\rho]$, where $\mathcal{A}_\rho$ holds the $\rho n$ positions with the highest KeyDiff score~\cite{keydiff} in document order, and the fraction $\rho$ is fixed in advance (by default, $\rho=r$).
Both methods thus fix how much to rehearse before reading the document.

\section{A Cache Keeps What It Rehearses}\label{sec:rehearsal}

In Eq.~\eqref{eq:score}, the rehearsal $R$ stands in for every future query: whatever $R$ contains is all the scorer knows about what will be asked.
We show that $R$ does more than inform the ranking; it decides where the budget goes.
All diagnostics in this section use the development split of RULER-4K~\cite{ruler} with Qwen3-8B at $r=5\%$ (13 tasks, 650 questions), disjoint from every test split in Section~\ref{sec:experiments}. %

\ifgeneral\else
\begin{figure}[t]
\centering
\includegraphics[width=\columnwidth]{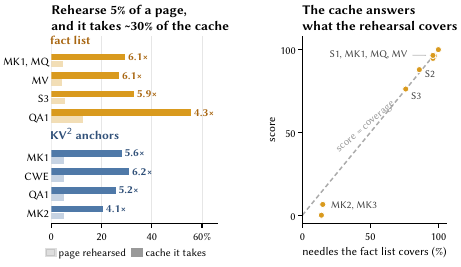}
\caption{\textbf{A cache keeps what it rehearses} (Qwen3-8B, RULER-4K development split, keep ratio 5\%). (\textbf{\textit{Left}}) Share of the document that a rehearsal covers (light) and share of the kept KV entries, over all layers and heads, that fall on those positions (dark), for a \www{}{semantic rehearsal (}fact list\www{}{)} and \www{}{a geometric one (}KV$^2$ anchors\www{}{)}. (\textbf{\textit{Right}}) Rehearsing only the fact list: needle coverage of the list against the score. MK/MQ/MV: multi-key/-query/-value needles; S: single needle; CWE: common words; QA: question answering.} %
\label{fig:enrich}
\end{figure}
\fi

\ifgeneral\else
\begin{figure*}[t]
  \centering
  \includegraphics[width=\textwidth]{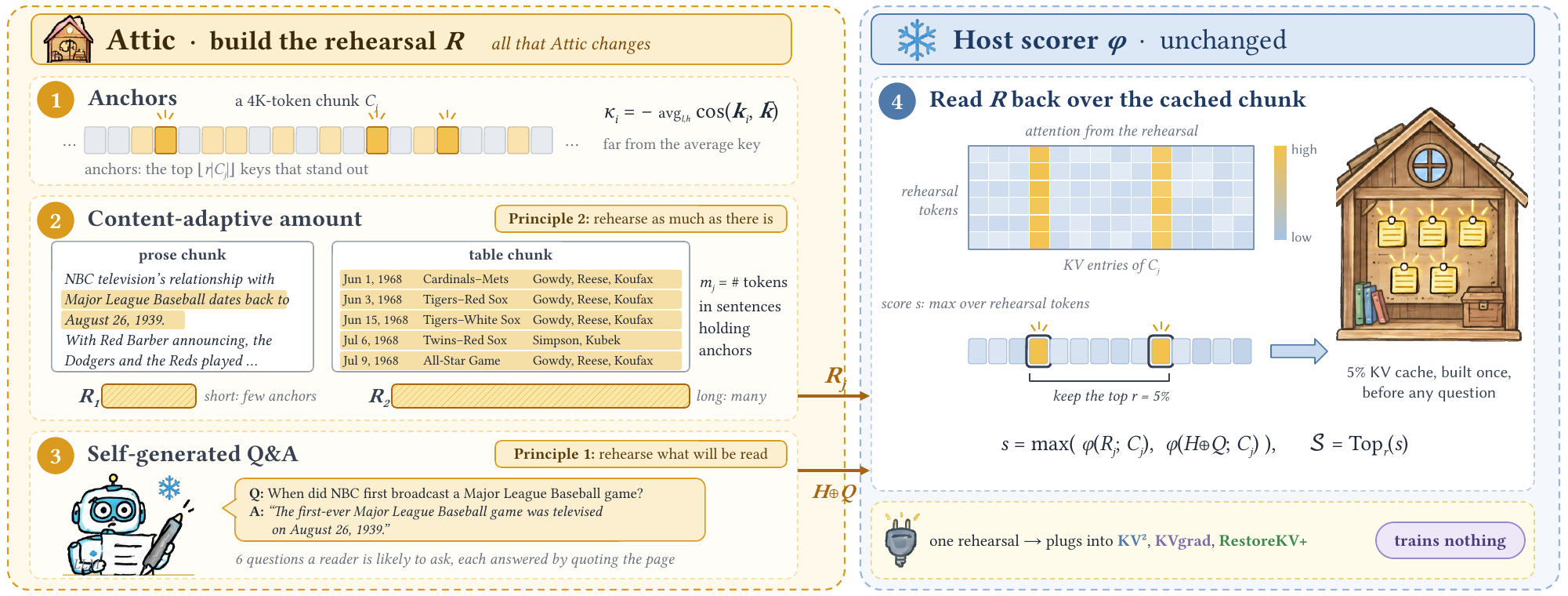}
  \caption{\textbf{Overview of \ourmethod.} (\textbf{\textit{Left}}) \ourmethod builds the rehearsal $R$, the only part of compression it changes: \ding{182} the keys that stand out from the chunk's average key are the anchors; \ding{183} the anchors are expanded to $m_j$ tokens, the length of the sentences that hold them, so a prose chunk is rehearsed briefly and a table of facts at length (Principle~2); \ding{184} the model writes the questions a reader is likely to ask and answers them by quoting the page (Principle~1). (\textbf{\textit{Right}}) The host's scorer $\phi$ reads the rehearsal back over the cached chunk, scores every KV entry by the attention it receives (heat map illustrative), and keeps the top $r$. \www{Text and Q\&A pair are from the LooGLE page of Figure~\ref{fig:teaser}; the same rehearsal plugs into KV$^2$, KVgrad and RestoreKV+.}{Text is from the LooGLE page of Figure~\ref{fig:teaser}; the Q\&A pair is the first one the model wrote for it. The same rehearsal plugs into KV$^2$, KVgrad and RestoreKV+, and nothing is trained.}} %
  \label{fig:framework}
\end{figure*}
\fi

\ifgeneral
\begin{table*}[t]
\caption{Main results on the test splits at a 5\% keep ratio (R-4K$^{3\%}$: 3\%). R: RULER; LB: LongBench (14 natural-text tasks). Reh.: rehearsal length, including QA pairs, as a fraction of the context (Qwen3-8B, RULER-4K test). \ourmethod{} alone uses KV$^2$'s scorer; a shaded row below a host adds \ourmethod's rehearsal to it; \gain{x}: gain over KV$^2$ or the host. KV$^2$ uses one $\rho$ per model and keep ratio, selected on the RULER-4K development split and used on every benchmark; \ourmethod{} uses one configuration throughout. $^\ddagger$Re-selects $\rho$ on LongBench's development split, which requires knowing in advance which benchmark a document comes from; not ranked. $^\dagger$Restore tokens released by the RestoreKV authors (none for Qwen3-4B/14B); all other rows are training-free. \textbf{Bold}: above every baseline; \underline{underlined}: the strongest baseline. Avg.\ Rank: mean rank over the columns each row has.}
\label{tab:main}
\footnotesize
\setlength{\tabcolsep}{3.6pt}
\setlength{\aboverulesep}{0pt}\setlength{\belowrulesep}{0pt}
\renewcommand{\arraystretch}{1.12}
\setlength{\gw}{2.45cm}
\begin{tabular}{p{\gw}l c *{8}{r} c}
\toprule
\headrow & & & \multicolumn{4}{c}{\textbf{Qwen3-8B}} & \multicolumn{2}{c}{\textbf{Llama-3.1-8B}} & \multicolumn{1}{c}{\textbf{Qwen3-4B}} & \multicolumn{1}{c}{\textbf{Qwen3-14B}} & \\
\headrow \textbf{Group} & \textbf{Method} & \textbf{Reh.} & \multicolumn{1}{c}{\textbf{R-4K}\hspace*{0.6\gainw}} & \multicolumn{1}{c}{\textbf{R-16K}\hspace*{0.6\gainw}} & \multicolumn{1}{c}{\textbf{R-4K$^{3\%}$}\hspace*{0.6\gainw}} & \multicolumn{1}{c}{\textbf{LB}\hspace*{0.6\gainw}} & \multicolumn{1}{c}{\textbf{R-4K}\hspace*{0.6\gainw}} & \multicolumn{1}{c}{\textbf{LB}\hspace*{0.6\gainw}} & \multicolumn{1}{c}{\textbf{R-4K}\hspace*{0.6\gainw}} & \multicolumn{1}{c}{\textbf{R-4K}\hspace*{0.6\gainw}} & \textbf{Avg.\ Rank}$\downarrow$ \\
\midrule
 & Full cache (reference) & -- & 96.5\gslot{} & 93.5\gslot{} & 96.5\gslot{} & 47.1\gslot{} & 95.8\gslot{} & 45.1\gslot{} & 94.6\gslot{} & 96.8\gslot{} & -- \\
\midrule
 & StreamingLLM~\cite{streamingllm} & 0.00 & 11.4\gslot{} & 12.9\gslot{} & 6.1\gslot{} & 28.1\gslot{} & 15.2\gslot{} & 29.3\gslot{} & 10.4\gslot{} & 20.8\gslot{} & 11.62 \\
 & SnapKV~\cite{snapkv} & 0.00 & 14.9\gslot{} & 17.8\gslot{} & 8.9\gslot{} & 32.2\gslot{} & 15.2\gslot{} & 29.0\gslot{} & 13.9\gslot{} & 19.1\gslot{} & 10.88 \\
 & TOVA~\cite{tova} & 0.00 & 6.9\gslot{} & 28.8\gslot{} & 3.4\gslot{} & 31.6\gslot{} & 14.3\gslot{} & 28.7\gslot{} & 4.8\gslot{} & 10.5\gslot{} & 12.75 \\
 & Expected Attention~\cite{expectedattention} & 0.00 & 9.2\gslot{} & 8.6\gslot{} & 7.8\gslot{} & 36.5\gslot{} & 17.1\gslot{} & 31.8\gslot{} & 10.0\gslot{} & 12.8\gslot{} & 11.00 \\
 & KeyDiff~\cite{keydiff} & 0.00 & 23.4\gslot{} & 32.2\gslot{} & 14.1\gslot{} & 19.1\gslot{} & 49.0\gslot{} & 28.8\gslot{} & 20.6\gslot{} & 33.4\gslot{} & 10.25 \\
\multirow{-6}{\gw}{\raggedright\itshape (a) Score, no rehearsal} & Compactor~\cite{compactor} & 0.00 & 35.3\gslot{} & 49.5\gslot{} & 25.5\gslot{} & 35.6\gslot{} & 50.8\gslot{} & 35.1\gslot{} & 30.3\gslot{} & 34.9\gslot{} & 7.62 \\
\midrule
\multirow{-1}{\gw}{\raggedright\itshape (b) Rehearse a task guide} & TaskPress~\cite{taskpress} & 0.02 & 34.0\gslot{} & 51.6\gslot{} & 27.3\gslot{} & 37.1\gslot{} & 47.2\gslot{} & 34.1\gslot{} & 30.6\gslot{} & 34.0\gslot{} & 7.75 \\
\midrule
\multirow{-1}{\gw}{\raggedright\itshape (c) Rehearse everything} & KVzip+~\cite{kvzap} & 1.00 & 58.6\gslot{} & 60.1\gslot{} & 31.5\gslot{} & 28.6\gslot{} & 66.2\gslot{} & 29.6\gslot{} & 56.5\gslot{} & 77.4\gslot{} & 7.50 \\
\midrule
 & KV$^2$~\cite{kvsquared} & 0.35 & 76.8\gslot{} & 74.5\gslot{} & \underline{67.0}\gslot{} & 38.6\gslot{} & 74.8\gslot{} & 35.0\gslot{} & \underline{74.1}\gslot{} & \underline{81.0}\gslot{} & 4.75 \\
\multirow{-2}{\gw}{\raggedright\itshape (d) Rehearse a fixed share} & \quad $\rho$ per benchmark$^\ddagger$ &  & \gslot{} & \gslot{} & \gslot{} & 39.1\gslot{} & \gslot{} & \underline{39.1}\gslot{} & \gslot{} & \gslot{} & -- \\
\midrule
\oursrow \multirow{-1}{\gw}{\raggedright\itshape Ours} & \ourmethod{} & 0.42 & \textbf{82.1}\gslot{\gain{5.3}} & \textbf{75.1}\gslot{\gain{0.6}} & \textbf{73.4}\gslot{\gain{6.4}} & \textbf{40.6}\gslot{\gain{2.0}} & 77.2\gslot{\gain{2.4}} & \textbf{39.3}\gslot{\gain{4.3}} & \textbf{76.5}\gslot{\gain{2.4}} & \textbf{87.6}\gslot{\gain{6.6}} & 2.62 \\
\midrule
 & KVgrad~\cite{kvgrad} & 1.00 & 72.0\gslot{} & \underline{74.7}\gslot{} & 53.5\gslot{} & \underline{40.3}\gslot{} & 77.1\gslot{} & 37.9\gslot{} & 72.1\gslot{} & 79.5\gslot{} & 4.50 \\
\oursrow \multirow{-2}{\gw}{\raggedright\itshape (e) Gradient scoring} & \quad + \ourmethod & 0.42 & \textbf{83.7}\gslot{\gain{11.7}} & \textbf{80.6}\gslot{\gain{5.9}} & \textbf{70.6}\gslot{\gain{17.1}} & \textbf{41.9}\gslot{\gain{1.6}} & 80.3\gslot{\gain{3.2}} & \textbf{39.3}\gslot{\gain{1.4}} & \textbf{81.4}\gslot{\gain{9.3}} & \textbf{88.2}\gslot{\gain{8.7}} & 1.88 \\
\midrule
 & RestoreKV+~\cite{restorekv}$^\dagger$ & 1.00 & \underline{78.2}\gslot{} & 73.3\gslot{} & 47.6\gslot{} & 33.1\gslot{} & \underline{85.6}\gslot{} & 37.6\gslot{} & --\gslot{} & --\gslot{} & 5.33 \\
\oursrow \multirow{-2}{\gw}{\raggedright\itshape (f) Trained tokens} & \quad + \ourmethod & 0.42 & \textbf{88.9}\gslot{\gain{10.7}} & \textbf{80.6}\gslot{\gain{7.3}} & \textbf{75.7}\gslot{\gain{28.1}} & \textbf{42.4}\gslot{\gain{9.3}} & \textbf{86.4}\gslot{\gain{0.8}} & \textbf{41.6}\gslot{\gain{4.0}} & --\gslot{} & --\gslot{} & 1.00 \\
\bottomrule
\end{tabular}
\end{table*}

\fi

\paragraph{Evidence I: rehearsed positions are enriched in the kept budget.}
\www{Let $P_R\subseteq[n]$ be the positions that a rehearsal $R$ reproduces. If $R$ merely informed the ranking, the share of the document it covers, $|P_R|/n$, and the share of kept entries in $P_R$ would be close; Figure~\ref{fig:enrich} shows them far apart for two rehearsals that share nothing but the scorer.}{For a rehearsal $R$, let $P_R\subseteq[n]$ be the document positions that $R$ reproduces.
We compare the share of the document that $R$ covers, $|P_R|/n$, with the share of the kept entries whose position lies in $P_R$.
If the rehearsal merely informed the ranking, the two shares would be close; Figure~\ref{fig:enrich} shows them far apart.
We test two rehearsals that share nothing but the scorer.}
The \textbf{fact list} is semantic: with the full cache, the model copies every sentence that states a specific fact, and only this list is rehearsed (its positions are matched back to the document by 4-gram overlap).
The \textbf{KV$^2$ anchors} are geometric: the tokens whose keys deviate most from the average key, chosen with no notion of a fact.
For both, on fact-sparse tasks, positions covering 4.4--5.6\% of the document attract 20.6--32.9\% of the kept budget, an enrichment of 4.1--6.2$\times$ and about 6$\times$ on most tasks. %
A rehearsal that covers more takes more: a fact list covering 12.8\% of a QA document draws 55.6\% of the budget. %
Whichever way the rehearsal is chosen, the budget flows to what is rehearsed.

\paragraph{Evidence II: a cache remembers what its rehearsal covers.}
If the budget follows the rehearsal, a cache should answer exactly the questions that its rehearsal covers.
Rehearsing only the fact list makes this testable, since we can count which needle sentences the list contains.
\www{}{The score follows the coverage task by task.
}\www{From MK1, MQ and MV to S2 and S3, the coverage falls from 96--97\% to 86\% and 76\%, and the score with it, from 94.7--96.5 to 87.9 and 76.2.}{The list covers 96--97\% of the needles on MK1, MQ, and MV, and the cache scores 94.7--96.5; it covers 86\% on S2 and scores 87.9, and 76\% on S3 and scores 76.2.} %
Where it misses most needles (14--15\% on MK2, MK3), it collapses to 0.0 and 6.5. %

Is the loss then a matter of capacity, a 5\% subset too small to hold what a question needs?
An answer-aware oracle settles this.
When the rehearsal is the question followed by its gold answer, attention scoring keeps a 5\% subset that scores 93.5, against 94.7 without compression. %
The subset has room.
Rehearsing the whole document instead, with the same scorer and budget, scores 37.8, and 58.9 even with the KVzip+ normalization of Eq.~\eqref{eq:score}: the loss comes from what is rehearsed, not from how much is kept. %

\paragraph{Evidence III: the answer is kept only where the rehearsal sends the budget.}
If the budget follows the rehearsal, it should reach the answer itself only when the rehearsal leads it there.
For each of the 425 questions on the eight needle tasks (Figure~\ref{fig:probe}), we measure the share of the answer's KV entries, over all layers and heads, that the cache keeps; at $r=5\%$, a cache that singles nothing out keeps about 5\%. %
Whole-document rehearsal keeps 5.7--7.1\% on seven of the eight tasks: the budget lands on the answer no more often than on any other token. %
The exception, S1, repeats one sentence as its haystack, so a reread reads back only the needle (50.1\%); in natural text nearly every token is unpredictable, and a full reread spreads the budget evenly. %
\www{}{The answer-aware oracle keeps 48.4\% on average.} %
\ourmethod, the rehearsal we build in Section~\ref{sec:method}, keeps 15.0--17.0\% on the five tasks that it lifts to 94--100 points, and stays at the 5\% share on MK2 and MK3, the two tasks it leaves unsolved. %
\ourmethod solves the tasks where it draws the budget to the answer, and only those.
At a 3\% keep ratio, whole-document rehearsal stays near chance (3.5--5.0\%), while \ourmethod still keeps 13.2--14.7\% of the answer on the same five tasks, now 4.5 times chance: a targeted rehearsal protects the answer as the budget shrinks, so the gap widens (Sec.~\ref{sec:rq2}). %

\paragraph{Rehearse everything, remember nothing.}
The three pieces of evidence identify the cause behind our headline: \textbf{a cache keeps what it rehearses}.
A full-document rehearsal singles out nothing, so it spreads the budget over the whole document, and most of \www{a Web document}{a document} is fluent prose.
Under a tight budget, that prose crowds out the few facts that will be asked.
The crowding sharpens as the budget shrinks, so the gap between full and targeted rehearsal must widen as $r$ falls; Section~\ref{sec:rq2} confirms this prediction.

\section{Two Principles and \ourmethod}\label{sec:method}

Section~\ref{sec:rehearsal} reduces query-agnostic compression to the design of the rehearsal $R$, which has two degrees of freedom: what it contains, and how much of it there is.
We state one principle for each, ground each in controlled experiments, and build \ourmethod from the two.

\ifgeneral\else

\fi

\subsection{Principle 1: Rehearse What Will Be Read}\label{sec:p1}
A future query reads out a few spans of a document: the value bound to a key, the date of an event, the name in a claim.
The rehearsal should reproduce these spans as they would be read out, not the document.
Three experiments support this principle.

\textbf{(I) Rehearsing only what is read is nearly lossless.}
On the needle tasks of RULER-4K, rehearsing only the needle sentences, i.e., every fact that can be asked, under the same budget as KVzip+ scores 98.5--100 on every fact-sparse task at both $r=5\%$ and $r=3.125\%$; its lead over KVzip+ grows from 24--58 to 74--83 points\www{}{(on MK1, KVzip+ falls from 44.4 to 18.5)}. %

\textbf{(II) Reading the question is not reading the answer.}
Rehearsing the question alone, without its answer, scores 47.0, against 93.5 with the answer under the same scorer. %
While reading a question, attention lands on the key it names; only while producing the answer does it land on the value.
A rehearsal must therefore reproduce the read-out, not the request.

\textbf{(III) Self-generated read-outs carry over to natural text.}
On LongBench~\cite{longbench}, an answer-aware rehearsal keeps a 5\% subset that matches the uncompressed model (44.9 vs.\ 44.4), so on natural text, too, the loss lies in what is rehearsed. %
A deployable rehearsal cannot see the questions, but the model can anticipate them: asked to write the questions a reader is likely to ask and to answer each by quoting the document, it produces read-outs of the spans it expects to be asked.
Adding this \textbf{self-generated QA rehearsal} to an anchor rehearsal raises the LongBench development average from 38.2 to 39.6; Section~\ref{sec:ablation} shows the same gain on every RULER test setting. %

\ifgeneral
\begin{table}[t]
\caption{\www{The Web document cache}{Compress once, answer many} on LooGLE (Qwen3-8B, keep ratio 5\%): each document is compressed once, then answers all of its questions (ROUGE-L / METEOR, $\times$100). The shaded row below RestoreKV+ adds \ourmethod's rehearsal to it; \gain{x}/\loss{x}: change over KV$^2$ or the host. $^\dagger$Released restore tokens. \textbf{Bold}: above every baseline; \underline{underlined}: the strongest baseline.}
\label{tab:loogle}
\footnotesize
\setlength{\tabcolsep}{4.2pt}
\setlength{\aboverulesep}{0pt}\setlength{\belowrulesep}{0pt}
\renewcommand{\arraystretch}{1.1}
\begin{tabular}{l rr rr}
\toprule
\headrow & \multicolumn{2}{c}{\textbf{Short dependency}} & \multicolumn{2}{c}{\textbf{Long dependency}} \\
\headrow \textbf{Method} & \multicolumn{1}{c}{\textbf{ROUGE-L}\hspace*{0.6\gainw}} & \multicolumn{1}{c}{\textbf{METEOR}\hspace*{0.6\gainw}} & \multicolumn{1}{c}{\textbf{ROUGE-L}\hspace*{0.6\gainw}} & \multicolumn{1}{c}{\textbf{METEOR}\hspace*{0.6\gainw}} \\
\midrule
Full cache & 65.7\gslot{} & 34.2\gslot{} & 31.9\gslot{} & 12.3\gslot{} \\
\midrule
 KVzip+ & 28.1\gslot{} & 14.0\gslot{} & 19.6\gslot{} & 6.4\gslot{} \\
 KV$^2$ & 34.3\gslot{} & 17.2\gslot{} & 23.6\gslot{} & \underline{7.8}\gslot{} \\
\oursrow \ourmethod & 36.2\gslot{\gain{1.9}} & 18.6\gslot{\gain{1.4}} & 22.9\gslot{\loss{0.7}} & 7.8\gslot{} \\
\midrule
 RestoreKV+$^\dagger$ & \underline{39.0}\gslot{} & \underline{19.3}\gslot{} & \underline{25.3}\gslot{} & 7.2\gslot{} \\
\oursrow \quad + \ourmethod & \textbf{40.9}\gslot{\gain{1.9}} & \textbf{20.7}\gslot{\gain{1.4}} & \textbf{26.1}\gslot{\gain{0.8}} & \textbf{9.2}\gslot{\gain{2.0}} \\
\bottomrule
\end{tabular}
\end{table}

\fi
\subsection{Principle 2: Rehearse As Much As There Is}\label{sec:p2}
Documents differ in how much of them can be asked.
A narrative may hide one fact in a page of prose, while a table or a list of records is facts from top to bottom.
The amount of rehearsal should follow this content instead of being fixed before the document is read.

\textbf{(I) A content-adaptive amount beats a fixed one at equal average.}
On the RULER-4K development split with Qwen3-8B, we compare two anchor rehearsals that rehearse about 35\% of the context on average.
KV$^2$ with $\rho=0.35$ rehearses 35\% of every document and scores 74.6; the content-adaptive amount of \ourmethod (Section~\ref{sec:attic}) rehearses 11--100\% per document, 34\% on average, and scores 78.0. %
The gain comes where the content is dense: on MK2 and MK3, whose contexts are lists of key--value statements, the score rises from 27.0 to 54.0 and from 4.3 to 28.3. %
\www{The amount tracks the content (Spearman correlation 0.59 with the number of fact statements, $n=425$): contexts with ten or more facts are rehearsed in full, those with at most one at a median of 28\%.}{The amount tracks the content: over the needle-task contexts it correlates with the number of fact statements (Spearman correlation 0.59, $n=425$), and contexts with ten or more facts are rehearsed in full (median 100\%), while those with at most one are rehearsed at a median of 28\%.} %
On Llama-3.1-8B~\cite{llama3}, it scores 76.1, against 73.9 for the best $\rho$. %

\textbf{(II) No fixed fraction is right.}
Tuning $\rho$ cannot close this gap: $\rho\in\{0.25,0.35,0.5\}$ gives 74.0, 74.6, and 73.9, all below 78.0. %
The best $\rho$ drifts with the benchmark (Section~\ref{sec:rq4}) and with the budget, as KV$^2$ itself observes~\cite{kvsquared}: on Qwen3-8B it is 0.25 at $r=3\%$, 0.35 at 5\%, and 0.9 from 10\% to 30\%. %
\www{}{A fraction fixed in advance is right for some documents, benchmarks, and budgets, and wrong for the rest.}

\textbf{(III) KVzip and KV$^2$ are fixed-amount special cases.}
KVzip rehearses everything and KV$^2$ a fixed fraction, whatever the document holds; both fix the amount, and differ only in where.
In \ourmethod, the number of anchors equals the keep ratio, so its rehearsal grows as the budget loosens and approaches the full document: KVzip is the loose-budget limit of Principle 2, not its opposite, and at loose budgets all methods meet (Section~\ref{sec:rq2}).

\subsection{\ourmethod}\label{sec:attic}
\ourmethod builds the rehearsal of Eq.~\eqref{eq:score} from the two principles: \textbf{anchors} decide what to rehearse, a \textbf{content-adaptive amount} decides how much, and a \textbf{self-generated QA rehearsal} adds the read-outs the model expects.
It changes only the rehearsal; the scorer and the selection $\mathrm{Top}_r$ stay untouched.
Figure~\ref{fig:framework} illustrates it\ifgeneral, and Algorithm~\ref{alg:attic} (Appendix~\ref{app:alg}) states it in full\fi.

\textbf{\ding{182} Anchors (what).}
As in KV$^2$, we split $C$ into chunks of 4,096 tokens. %
Each position $i$ receives the KeyDiff score~\cite{keydiff} $\kappa_i=-\frac{1}{H}\sum_{l,h}\cos(\mathbf{k}^{l,h}_i,\bar{\mathbf{k}}^{l,h})$, where $\bar{\mathbf{k}}^{l,h}$ is the mean normalized key of the document, so that tokens whose keys stand out from the average key score high.
The anchors $\mathcal{A}_j$ of chunk $C_j$ are its $\lfloor r|C_j|\rfloor$ highest-scoring positions.
The anchor ratio equals the keep ratio, so \ourmethod adds no ratio to tune.

\textbf{\ding{183} Content-adaptive amount (how much).}
We segment $C_j$ into sentences, ending a sentence at any token that contains a period, question mark, exclamation mark, or line break, and let $m_j$ be the number of tokens in the sentences that contain at least one anchor.
The anchor rehearsal $R_j$ consists of the $m_j$ highest-scoring positions of $C_j$ under $\kappa$, in document order: the anchors, expanded to $m_j$ tokens.
A fact-dense chunk spreads its anchors over many sentences and receives a large $m_j$; a chunk of prose concentrates them in a few and receives a small one.
\www{Sentences are a measuring stick, not the unit: rehearsing the sentences themselves adds 0.6 points on Qwen3-8B and costs 4.4 on Llama-3.1-8B.}{Sentences are a measuring stick, not the unit of rehearsal. Rehearsing the sentences themselves instead adds only 0.6 points on Qwen3-8B (78.6 vs.\ 78.0) and costs 4.4 on Llama-3.1-8B (71.7 vs.\ 76.1): the gain comes from the amount, not from the unit.} %
\www{The positions are rehearsed in a single pass in document order, so the rehearsal continues itself from local context and reads back mainly what it cannot predict, the needles, entities and numbers that queries ask for; interleaved sparse passes drop the RULER-4K development score from 81.8 to 69.8 (Appendix~\ref{app:coherent}).}{The positions are rehearsed in a single pass in document order, which lets the rehearsal continue itself from local context, so that it reads back from the document mainly what it cannot predict, the needles, entities and numbers that queries ask for. Rehearsing the same positions as interleaved sparse passes drops the RULER-4K development score from 81.8 to 69.8 (Appendix~\ref{app:coherent}).} %

\textbf{\ding{184} Self-generated QA rehearsal (read-outs).}
Before compression, the model reads the document with its full cache and generates at most 256 tokens: 6 questions a reader is likely to ask, each answered by quoting the exact words of the text (prompt in Appendix~\ref{app:prompt}). %
The output $Q$ is rehearsed behind a one-line header $\textsc{H}$ against every chunk.
Because the answers quote the document, rehearsing them reproduces the read-out of the quoted spans (Section~\ref{sec:p1}, II).
Writing $\phi(R;C_j)$ for the scores that a rehearsal $R$ gives to the entries of chunk $C_j$, the final score on that chunk is the entry-wise maximum $s=\max\big(\phi(R_j;C_j),\,\phi(\textsc{H}\oplus Q;C_j)\big)$, and $\mathcal{S}=\mathrm{Top}_r(s)$.

\textbf{\ding{185} Restore tokens (optional).}
RestoreKV~\cite{restorekv} acts on the capacity side: 8 learned restore tokens and a LoRA adapter help the model read from a sparse cache, and the selection is left untouched. %
Since \ourmethod acts on the selection side, the two compose directly: we attach the officially released restore weights, trained on KVzip+ selections, to the cache that \ourmethod selects, and train nothing.

\paragraph{\ourmethod as a plug-in.}
A rehearsal-scoring \textit{host method} is a pair of a rehearsal and a scoring rule $\phi$, where Eq.~\eqref{eq:score} is one choice of $\phi$.
``Host + \ourmethod'' keeps the host's $\phi$ and replaces its rehearsal with that of \ourmethod.
KV$^2$ is the special case of \ourmethod with a fixed amount and no QA rehearsal; \ourmethod shares its KeyDiff anchors and scorer and sets the content and amount of the rehearsal by the two principles.
KVgrad + \ourmethod keeps the gradient attribution of KVgrad~\cite{kvgrad}\www{}{ (attention weight $\times$ output-contribution norm $\times$ a gradient norm computed on the rehearsal, originally the full document in 256-token pieces)} and applies it to $R_j$ and $\textsc{H}\oplus Q$ in pieces of at most 256 tokens, combined by the entry-wise maximum. %

\section{Experiments}\label{sec:experiments}

\www{We ask whether \ourmethod outperforms existing query-agnostic methods, alone and inside their hosts (\textbf{RQ1}); how its advantage changes with the keep ratio (\textbf{RQ2}); whether it carries over to the Web document cache, and at what cost (\textbf{RQ3}); and whether a tuned fixed ratio can catch up (\textbf{RQ4}); then we ablate it.}{We evaluate \ourmethod through four research questions.
\textbf{RQ1}: Does \ourmethod outperform existing query-agnostic methods, both on its own and when plugged into their host methods?
\textbf{RQ2}: How does its advantage change with the keep ratio?
\textbf{RQ3}: Does it carry over to \www{the Web document cache}{compress-once, answer-many caching}, where a \www{page}{document} is compressed once and then answers all of its questions, and what does the compression cost?
\textbf{RQ4}: Can a fixed rehearsal ratio, however well tuned, catch up with \ourmethod?
We close with an ablation and a per-category view of LongBench.}

\subsection{Experimental Setup}\label{sec:exp-setup}

\paragraph{\textbf{Models and benchmarks.}}
We use four open LLMs from two families: Qwen3-8B, Qwen3-4B and Qwen3-14B~\cite{qwen3}, and Llama-3.1-8B-Instruct~\cite{llama3}.
We evaluate on three long-context benchmarks:
\textbf{(1) RULER}~\cite{ruler} at 4K and 16K context, whose 13 subtasks cover needle retrieval, multi-hop tracing, aggregation and question answering; %
\textbf{(2) LongBench}~\cite{longbench}, whose English tasks include multi-hop QA over Wikipedia pages, QA over scientific papers, and summarization of news and reports; we report the average over its 14 natural-text tasks\footnote{We leave out LongBench's Synthetic category, passage\_count and passage\_retrieval\_en, which count or index paragraphs rather than ask about their content; 16-task results are in Appendix~\ref{app:longbench}.}; %
and \textbf{(3) LooGLE}~\cite{loogle}, long Wikipedia pages and arXiv papers, each paired with many questions, which we use for RQ3.
The main keep ratio is 5\%; we add 3\% on Qwen3-8B RULER-4K and sweep 3--30\% in RQ2.
All choices are made on development splits of RULER and LongBench, and all reported numbers come from disjoint test splits (Appendix~\ref{app:impl}); LooGLE is used only for testing.

\paragraph{\textbf{Baselines.}}
We compare against six families of query-agnostic methods, grouped as in Table~\ref{tab:main}.
\textbf{(a) Scoring without rehearsal}: StreamingLLM~\cite{streamingllm}, SnapKV~\cite{snapkv}, TOVA~\cite{tova}, Expected Attention~\cite{expectedattention}, KeyDiff~\cite{keydiff} and Compactor~\cite{compactor}; kvpress compresses the context before the question is appended, so each of them runs query-agnostic by construction.
For rehearsal scoring we take the strongest representative of each line:
\textbf{(b) rehearse a task guide}, TaskPress~\cite{taskpress}, \www{given an LLM-written task description;}{which is given an LLM-written description of each task (one for RULER, one per LongBench task), information a query-agnostic cache does not have;} %
\textbf{(c) rehearse everything}, KVzip+~\cite{kvzap};
\textbf{(d) rehearse a fixed share}, KV$^2$~\cite{kvsquared};
\textbf{(e) gradient scoring}, KVgrad~\cite{kvgrad}, which scores the same full rehearsal by gradient attribution;
and \textbf{(f) trained tokens}, RestoreKV+~\cite{restorekv}, which adds restore tokens trained by its authors to the KVzip+ selection.
\www{We run every method in kvpress~\cite{kvpress} with the same models, splits and prompts. KV$^2$'s $\rho$ is selected per model and keep ratio on the RULER-4K development split and used on every benchmark (Appendix~\ref{app:rho}).}{We run every method ourselves in kvpress~\cite{kvpress}, with the same models, splits and prompts.
For KV$^2$, $\rho$ is selected on the RULER-4K development split for each model and keep ratio and used on every benchmark, as \ourmethod uses one configuration everywhere; Table~\ref{tab:main} also reports it with $\rho$ re-selected on LongBench ($\ddagger$; Appendix~\ref{app:rho}).}

\paragraph{\textbf{Configurations of \ourmethod.}}
\ourmethod uses one configuration for all models, benchmarks and budgets: the anchor ratio equals the keep ratio, and the model writes 6 question--answer pairs of at most 256 tokens. These constants were fixed once on the development splits with Qwen3-8B and are never re-selected. %
``+ \ourmethod'' rows follow Section~\ref{sec:attic}; restore tokens exist only for Qwen3-8B and Llama.

\ifgeneral\else
\begin{table}[t]
\caption{\www{The Web document cache}{Compress once, answer many} on LooGLE (Qwen3-8B, keep ratio 5\%): each document is compressed once, then answers all of its questions (ROUGE-L / METEOR, $\times$100). The shaded row below RestoreKV+ adds \ourmethod's rehearsal to it; \gain{x}/\loss{x}: change over KV$^2$ or the host. $^\dagger$Released restore tokens. \textbf{Bold}: above every baseline; \underline{underlined}: the strongest baseline.}
\label{tab:loogle}
\footnotesize
\setlength{\tabcolsep}{4.2pt}
\setlength{\aboverulesep}{0pt}\setlength{\belowrulesep}{0pt}
\renewcommand{\arraystretch}{1.1}
\begin{tabular}{l rr rr}
\toprule
\headrow & \multicolumn{2}{c}{\textbf{Short dependency}} & \multicolumn{2}{c}{\textbf{Long dependency}} \\
\headrow \textbf{Method} & \multicolumn{1}{c}{\textbf{ROUGE-L}\hspace*{0.6\gainw}} & \multicolumn{1}{c}{\textbf{METEOR}\hspace*{0.6\gainw}} & \multicolumn{1}{c}{\textbf{ROUGE-L}\hspace*{0.6\gainw}} & \multicolumn{1}{c}{\textbf{METEOR}\hspace*{0.6\gainw}} \\
\midrule
Full cache & 65.7\gslot{} & 34.2\gslot{} & 31.9\gslot{} & 12.3\gslot{} \\
\midrule
 KVzip+ & 28.1\gslot{} & 14.0\gslot{} & 19.6\gslot{} & 6.4\gslot{} \\
 KV$^2$ & 34.3\gslot{} & 17.2\gslot{} & 23.6\gslot{} & \underline{7.8}\gslot{} \\
\oursrow \ourmethod & 36.2\gslot{\gain{1.9}} & 18.6\gslot{\gain{1.4}} & 22.9\gslot{\loss{0.7}} & 7.8\gslot{} \\
\midrule
 RestoreKV+$^\dagger$ & \underline{39.0}\gslot{} & \underline{19.3}\gslot{} & \underline{25.3}\gslot{} & 7.2\gslot{} \\
\oursrow \quad + \ourmethod & \textbf{40.9}\gslot{\gain{1.9}} & \textbf{20.7}\gslot{\gain{1.4}} & \textbf{26.1}\gslot{\gain{0.8}} & \textbf{9.2}\gslot{\gain{2.0}} \\
\bottomrule
\end{tabular}
\end{table}

\fi

\ifgeneral
\begin{figure}[t]
\centering
\includegraphics[width=\columnwidth]{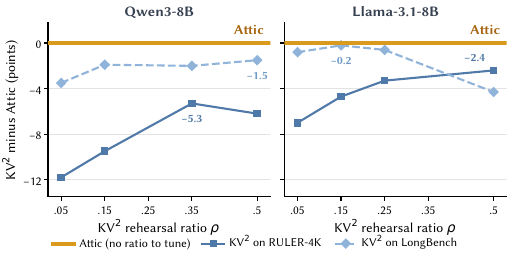}
\caption{Tuning a fixed rehearsal ratio, even on the test split, cannot catch up with \ourmethod (keep ratio 5\%). Each line is KV$^2$'s score minus \ourmethod's on the same benchmark, so \ourmethod is the zero line. Every point lies below it: no single $\rho$ gets close to \ourmethod on both benchmarks at once.}
\label{fig:rho}
\end{figure}
\fi

\subsection{Main Results (RQ1)}\label{sec:rq1}

Table~\ref{tab:main} covers seven model--benchmark pairs and a 3\% budget.

\textbf{Obs.\ding{182} Rehearsing nothing and rehearsing everything both collapse at \www{Web-scale}{tight} budgets; \ourmethod is the best training-free method in all eight settings.}
Scoring without rehearsal fails first: the six heuristics score 3.4--50.8 on RULER. KeyDiff, whose score picks \ourmethod's anchors, keeps 23.4 on Qwen3-8B RULER-4K; rehearsing those anchors instead of keeping them reaches 82.1. %
Rehearsing everything is no cure: KVzip+ keeps 58.6 of the 96.5 uncompressed score on Qwen3-8B RULER-4K and only 31.5 at 3\%, and on Qwen3-8B LongBench four of the six heuristics, which rehearse nothing, score above it (31.6--36.5 vs.\ 28.6). %
Knowing the task does not close the gap either: TaskPress, given a description of what each task asks, scores 27.3--51.6 on RULER and stays 3.5 and 5.2 points below \ourmethod on the two LongBench columns. %
\ourmethod scores 82.1 and 73.4 in the two RULER settings, 5.3 and 6.4 points above the strongest training-free baseline, KV$^2$ with a dev-tuned $\rho$, and on Qwen3-8B at 5\% it also exceeds the trained RestoreKV+ (78.2). %
It leads every training-free baseline across context length (up to 64K tokens, Appendix~\ref{app:long}), model size and natural text, by up to 6.6 points (87.6 vs.\ 81.0 on Qwen3-14B). %
Where a baseline comes close, it is a host that \ourmethod improves: KVgrad is the nearest on RULER-16K, Qwen3-8B LongBench and Llama RULER-4K, and \ourmethod plugged into it adds 5.9, 1.6 and 3.2 points there (Obs.\ding{183}). %
KV$^2$ needs a different ratio for each benchmark: with the ratio it selects on RULER, it trails \ourmethod by 2.0 and 4.3 points on LongBench, and even $\rho$ re-selected on LongBench itself ($\ddagger$), which requires knowing each document's benchmark in advance, reaches only 39.1 (Section~\ref{sec:rq4}). %

\textbf{Obs.\ding{183} Plugged into a host, \ourmethod lifts it above every existing method that trains as little as it does.}
Plugged into KVgrad, the gradient-based host, \ourmethod raises it by 11.7 points on Qwen3-8B RULER-4K, 17.1 at the 3\% keep ratio, 3.2 on Llama, 9.3 on Qwen3-4B and 8.7 on Qwen3-14B; KVgrad + \ourmethod beats every existing training-free method in all eight settings, and on Qwen3-4B and Qwen3-14B (81.4 and 88.2) it is the best method overall. %
Plugged into RestoreKV+, the trained host, it raises it in every setting, from 0.8 points on Llama RULER-4K to 28.1 at the 3\% keep ratio, and RestoreKV+ + \ourmethod beats every existing method in all six settings where restore tokens exist, with an average rank of 1.00. %
\www{The two hosts differ in scorer and in training, yet both improve: full-context rehearsal fails by its target, not by its scorer or capacity.}{The two hosts differ in how they score a rehearsal (attention vs.\ gradient attribution) and in whether anything is trained, yet both improve once they rehearse what will be read, as much as there is.
Full-context rehearsal fails by its target, not by its scorer or capacity.}

\subsection{Effect of the Budget (RQ2)}\label{sec:rq2}

Our mechanism makes a quantitative prediction: if prose crowds facts out of a full-context rehearsal, the crowding, and with it our advantage, must grow as the budget shrinks and vanish once the budget holds everything rehearsed.
Figure~\ref{fig:teaser} (right) sweeps the keep ratio on Qwen3-8B RULER-4K; Table~\ref{tab:app-budget} extends the sweep to 30\%.

\textbf{Obs.\ding{184} The advantage follows a dose--response curve: the tighter the budget, the wider the gap.}
The gap between \ourmethod and KVzip+ is 1.1 points at a 10\% keep ratio, 23.5 at 5\% and 41.9 at 3\%; with restore tokens, the gap to RestoreKV+ grows from 0.3 to 10.7 to 28.1 over the same ratios. %
\www{}{At 10\% both of our rows still lead their lines (94.1 vs.\ 93.0 for KVzip+ and 92.3 for KV$^2$; 95.2 vs.\ 94.9 for RestoreKV+).} %
At 20\% and 30\%, every method lies within 0.7 points of the uncompressed 96.5, the loose-budget limit of Section~\ref{sec:p2}. %
\www{}{The regime where compression matters most, 3--5\%, is exactly where the gap is largest.}

\subsection{\www{The Web Document Cache}{Compress Once, Answer Many} (RQ3)}\label{sec:rq3}

LooGLE reproduces \www{the Web document cache}{this setting} directly: each long document is compressed once, before any question, and then answers all of its questions, 1951 over 105 documents in the short-dependency set and 459 over 60 in the long-dependency set. %

\textbf{Obs.\ding{185} Compressed once per page, RestoreKV+ + \ourmethod serves all of the page's questions best, and \ourmethod's compression cost grows more slowly with the page than full rehearsal's.}
RestoreKV+ + \ourmethod is the best compressed method on both sets and both metrics (Table~\ref{tab:loogle}), above RestoreKV+ by 1.9 and 0.8 ROUGE-L and by 1.4 and 2.0 METEOR\www{}{ on the short- and long-dependency sets}. %
On the short-dependency set, \ourmethod alone leads all training-free methods, 1.9 ROUGE-L above KV$^2$ and 8.1 above KVzip+. %
\www{}{Full-context rehearsal again loses most: KVzip+ keeps 28.1 of the uncompressed 65.7 ROUGE-L on short-dependency questions.} %

\paragraph{\textbf{Compression cost.}}
A page is compressed once, off the query path. Rehearsing only part of the page, \ourmethod compresses it without the QA pairs in 24--45\% less time than KVzip+ at every length we time (4K--64K tokens\www{}{; Qwen3-8B, and Llama-3.1-8B at 64K}); the QA pairs add one decode of at most 256 tokens, which this saving overtakes as the page grows. %
With the QA pairs decoded from the prefill that compression already computes, a 32K-token page compresses as fast as with KVzip+ (24.0 vs.\ 23.9\,s, about four prefills, repaid by four hits) and a 64K-token page 27\% faster (40.2 vs.\ 54.9\,s)\www{.}{; KVgrad runs out of memory from 32K tokens on.} %
\www{\paragraph{\textbf{What the budget buys.}}
Stored per head, as in Ada-KV~\cite{adakv}, a compressed page occupies exactly its keep ratio: over ten LooGLE pages of 20.5K tokens on average (Qwen3-8B; Figure~\ref{fig:webcache}a shows one), the KV of a page shrinks from 3.02\,GB to 151\,MB at 5\% and 91\,MB at 3\%. %
A hit skips the page's prefill: loading the compressed cache from host memory and reading the question takes 39\,ms against 3.29\,s to prefill the page again, 62--121 times faster per page, and loading it from disk is still 4.5--5.7 times faster (Figure~\ref{fig:webcache}b). %
With a Zipf(0.8) popularity over ten million 32K-token pages~\cite{breslau1999web}, a 1\,TB cache of full pages serves 8.5\% of requests; at 5\% and 3\% it serves 18.6\% and 20.9\%. %
Weighted by RULER-4K accuracy, full pages answer 8.2 of 100 requests correctly from the cache; KVzip+ and KV$^2$ peak at 10\% (14.6 and 14.5), while \ourmethod keeps rising, to 15.2 at 5\% and 15.4 at 3\%, where it leads KV$^2$ by 5.3 and 6.4 points (Figure~\ref{fig:webcache}c). %
}{Appendix~\ref{app:webcache} measures what a compressed page saves in a Web document cache: 1/20 of the storage at 5\%, and a first token in 39\,ms instead of 3.3\,s of prefill on a hit.} %
\ifgeneral\else
\begin{figure*}[t]
\centering
\includegraphics[width=\textwidth]{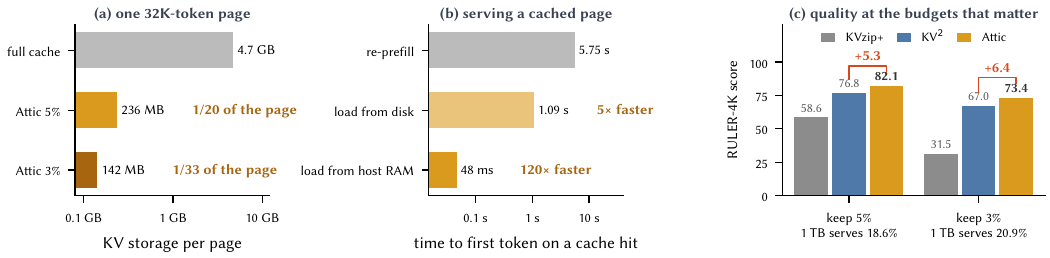}
\caption{\textbf{What a compressed Web document cache buys} (Qwen3-8B, one GPU). (a) KV storage of one 32K-token LooGLE page, stored per head. (b) Time to the first token on a cache hit, against prefilling the page again. (c) RULER-4K score at the keep ratios a Web-scale cache needs; below, the share of requests a 1\,TB cache serves (Zipf(0.8) over ten million 32K-token pages).}
\label{fig:webcache}
\end{figure*}
\fi

\ifgeneral\else
\begin{figure}[t]
\centering
\includegraphics[width=\columnwidth]{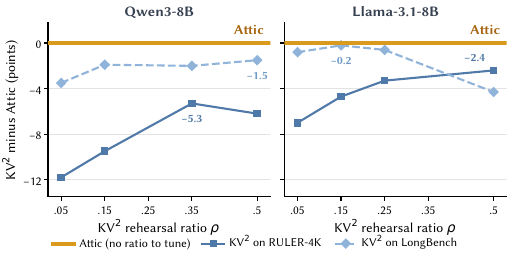}
\caption{Tuning a fixed rehearsal ratio, even on the test split, cannot catch up with \ourmethod (keep ratio 5\%). Each line is KV$^2$'s score minus \ourmethod's; every point lies below zero.}
\label{fig:rho}
\end{figure}
\fi

\subsection{Can a Tuned Ratio Catch Up? (RQ4)}\label{sec:rq4}

KV$^2$ rehearses a fixed fraction $\rho$ of each context; \ourmethod lets the content decide.
If, as \textit{rehearse as much as there is} states, the right amount depends on the content, the ratio KV$^2$ needs should move with it, and no single $\rho$ should match \ourmethod everywhere.

\textbf{Obs.\ding{186} The ratio KV$^2$ needs is a property of the content; \ourmethod reads it off the document.}
Selected on each development split, KV$^2$'s ratio for Llama is 0.5 on RULER-4K and 0.15 on LongBench (Qwen3-8B: 0.35 and 0.5). %
\www{}{A cache that holds both kinds of documents would have to know in advance which kind each one is; the content-adaptive amount reads this off each document.}
Figure~\ref{fig:rho} scans $\rho$ on the test splits, which gives KV$^2$ its best ratio in hindsight. On Llama, the best ratio for RULER-4K (0.5, 74.8) is the worst for LongBench (35.0), and no ratio comes within a point of \ourmethod on both (77.2 and 39.3); the single ratio that is best in hindsight still trails \ourmethod by 3.7 (Qwen3-8B) and 1.9 (Llama) points averaged over the two benchmarks. %
\www{}{The amount alone, without the QA rehearsal, already does what no ratio does: it beats every $\rho$ on RULER-4K (78.8 vs.\ at most 76.8 on Qwen3-8B, 75.6 vs.\ 74.8 on Llama; Table~\ref{tab:app-rhoscan}), and every $\rho$ that edges it on LongBench, by at most 0.8 points, falls 1.7--6.2 points behind it on RULER-4K.} %

\begin{table}[t]
\caption{Ablation on RULER test splits (keep ratio 5\% unless marked). Each block starts from a baseline and changes one thing per row; \gain{x}/\loss{x}: change from the row above. ``--'': no released restore tokens.}
\label{tab:ablation}
\footnotesize
\setlength{\tabcolsep}{2.4pt}
\setlength{\aboverulesep}{0pt}\setlength{\belowrulesep}{0pt}
\renewcommand{\arraystretch}{1.1}
\begin{tabular}{l rrr r r}
\toprule
\headrow & \multicolumn{3}{c}{\textbf{Qwen3-8B}} & \multicolumn{1}{c}{\textbf{Llama}} & \multicolumn{1}{c}{\textbf{Qwen3-4B}} \\
\headrow \textbf{Rehearsal} & \multicolumn{1}{c}{\textbf{4K}\hspace*{0.6\gainw}} & \multicolumn{1}{c}{\textbf{16K}\hspace*{0.6\gainw}} & \multicolumn{1}{c}{\textbf{4K$^{3\%}$}\hspace*{0.6\gainw}} & \multicolumn{1}{c}{\textbf{4K}\hspace*{0.6\gainw}} & \multicolumn{1}{c}{\textbf{4K}\hspace*{0.6\gainw}} \\
\midrule
\multicolumn{6}{l}{\textit{Rehearsal target, with the same self-generated QA}} \\
KVzip+: full rehearsal & 58.6\gslot{} & 60.1\gslot{} & 31.5\gslot{} & 66.2\gslot{} & 56.5\gslot{} \\ %
\quad + self-gen.\ QA & 65.8\gslot{\gain{7.2}} & 64.9\gslot{\gain{4.8}} & 35.4\gslot{\gain{3.9}} & 68.4\gslot{\gain{2.2}} & 61.7\gslot{\gain{5.2}} \\ %
\oursrow \quad full $\to$ anchors (= \ourmethod) & 82.1\gslot{\gain{16.3}} & 75.1\gslot{\gain{10.2}} & 73.4\gslot{\gain{38.0}} & 77.2\gslot{\gain{8.8}} & 76.5\gslot{\gain{14.8}} \\ %
\midrule
\multicolumn{6}{l}{\textit{Native KV pairs}} \\
KV$^2$: anchors, tuned fixed $\rho$ & 76.8\gslot{} & 74.5\gslot{} & 67.0\gslot{} & 74.8\gslot{} & 74.1\gslot{} \\ %
\quad + self-gen.\ QA & 78.1\gslot{\gain{1.3}} & 75.8\gslot{\gain{1.3}} & 72.2\gslot{\gain{5.2}} & 75.2\gslot{\gain{0.4}} & 75.0\gslot{\gain{0.9}} \\ %
\oursrow \quad $\rho \to$ adaptive amount (= \ourmethod) & 82.1\gslot{\gain{4.0}} & 75.1\gslot{\loss{0.7}} & 73.4\gslot{\gain{1.2}} & 77.2\gslot{\gain{2.0}} & 76.5\gslot{\gain{1.5}} \\ %
\midrule
\multicolumn{6}{l}{\textit{With released restore tokens}} \\
RestoreKV+: full rehearsal & 78.2\gslot{} & 73.3\gslot{} & 47.6\gslot{} & 85.6\gslot{} & --\gslot{} \\ %
\quad anchors, adaptive amount & 85.9\gslot{\gain{7.7}} & 78.8\gslot{\gain{5.5}} & 70.6\gslot{\gain{23.0}} & 85.3\gslot{\loss{0.3}} & --\gslot{} \\ %
\oursrow \quad + self-gen.\ QA & 88.9\gslot{\gain{3.0}} & 80.6\gslot{\gain{1.8}} & 75.7\gslot{\gain{5.1}} & 86.4\gslot{\gain{1.1}} & --\gslot{} \\ %
\bottomrule
\end{tabular}
\end{table}

\ifgeneral\edef\lbcatbegin{\noexpand\begin{table}[\lbcatpos]}\lbcatbegin
\caption{LongBench test results by official category (keep ratio 5\%). SD/MD: single-/multi-document QA; Sum.: summarization; FS: few-shot; Avg.: the 14 natural-text tasks, as in Table~\ref{tab:main}. Shaded rows below a host add \ourmethod's rehearsal to it. KV$^2$ uses the $\rho$ selected on RULER, as in Table~\ref{tab:main}; $^\ddagger$re-selects it on LongBench. $^\dagger$Released restore tokens. Per model, \textbf{bold}: above every baseline; \underline{underlined}: the strongest baseline.}
\label{tab:lbcat}
\footnotesize
\setlength{\tabcolsep}{3.4pt}
\setlength{\aboverulesep}{0pt}\setlength{\belowrulesep}{0pt}
\renewcommand{\arraystretch}{1.1}
\begin{tabular}{l cccccc}
\toprule
\headrow \textbf{Method} & \textbf{SD} & \textbf{MD} & \textbf{Sum.} & \textbf{FS} & \textbf{Code} & \textbf{Avg.} \\
\midrule
\headrow \multicolumn{7}{l}{\textit{Qwen3-8B} \quad (full cache: avg.\ 47.1)} \\
 KVzip+ & 23.8 & 16.3 & 18.5 & 46.3 & 42.9 & 28.6 \\
 KV$^2$ & 34.5 & 35.6 & 23.1 & 53.5 & 50.4 & 38.6 \\
 \quad $\rho$ per benchmark$^\ddagger$ & \underline{35.4} & 35.1 & 22.7 & 55.6 & 50.5 & 39.1 \\
\oursrow \ourmethod & \textbf{37.2} & 39.4 & 23.3 & 54.6 & 52.3 & \textbf{40.6} \\
 KVgrad & 31.0 & \underline{41.7} & 22.0 & \underline{57.6} & \underline{53.4} & \underline{40.3} \\
\oursrow \quad + \ourmethod & 32.1 & 41.5 & \textbf{24.0} & \textbf{60.4} & \textbf{56.4} & \textbf{41.9} \\
 RestoreKV+$^\dagger$ & 27.0 & 19.2 & \underline{23.4} & 51.7 & 49.8 & 33.1 \\
\oursrow \quad + \ourmethod & \textbf{38.3} & 39.1 & \textbf{25.0} & \textbf{57.9} & \textbf{56.2} & \textbf{42.4} \\
\midrule
\headrow \multicolumn{7}{l}{\textit{Llama-3.1-8B} \quad (full cache: avg.\ 45.1)} \\
 KVzip+ & 29.0 & 21.0 & 17.2 & 43.1 & 41.5 & 29.6 \\
 KV$^2$ & 34.6 & 31.7 & 22.6 & 46.7 & 41.5 & 35.0 \\
 \quad $\rho$ per benchmark$^\ddagger$ & \underline{37.2} & \underline{38.1} & 24.8 & 52.4 & \underline{45.1} & \underline{39.1} \\
\oursrow \ourmethod & \textbf{38.1} & \textbf{40.0} & 24.7 & 50.6 & 44.9 & \textbf{39.3} \\
 KVgrad & 33.2 & 34.2 & 23.7 & \underline{56.7} & 43.8 & 37.9 \\
\oursrow \quad + \ourmethod & \textbf{39.0} & 36.9 & 25.1 & 53.9 & 43.0 & \textbf{39.3} \\
 RestoreKV+$^\dagger$ & \underline{37.2} & 34.7 & \underline{25.4} & 52.6 & 38.4 & 37.6 \\
\oursrow \quad + \ourmethod & \textbf{43.0} & \textbf{43.6} & \textbf{26.3} & 54.8 & 39.8 & \textbf{41.6} \\
\bottomrule
\end{tabular}
\end{table}
\fi

\subsection{Ablation and Task Categories}\label{sec:ablation}

\textbf{Obs.\ding{187} Self-generated QA cannot rescue a full rehearsal; on anchors, the content-adaptive amount replaces a tuned ratio, and QA adds most where the budget is tightest.}
Table~\ref{tab:ablation} builds \ourmethod one step at a time from three baselines.
\www{Adding the same QA to a full-context rehearsal recovers only 2.2--7.2 points and stays 8.8--38.0 below \ourmethod, and 30.0 vs.\ 40.6 on Qwen3-8B LongBench: the full rehearsal crowds the answers out. Where the anchors sit matters too: at the same amount, random positions lose 3.2--20.0 points on RULER.}{The first block isolates the rehearsal target. Rehearsing the same self-generated QA on top of a full-context rehearsal recovers only 2.2--7.2 points and leaves it 8.8--38.0 below \ourmethod: as long as the whole context is rehearsed, it crowds the answers out of the budget, and only replacing it with anchors lets the QA pairs keep them. The same holds on natural text: on Qwen3-8B LongBench, full rehearsal with the QA pairs scores 30.0 against 40.6 for \ourmethod. Where the anchors sit matters as well: rehearsing the same amount at random positions instead of the KeyDiff anchors, with the same QA pairs, loses 3.2--20.0 points on the five RULER settings.} %
On KV$^2$'s anchors, self-generated QA adds 0.4 to 5.2 points, most at 3\%; with the QA in place, replacing the dev-tuned ratio with the content-adaptive amount adds a further 1.6 points on average and up to 4.0, with nothing tuned.\www{}{ Replacing the full rehearsal of RestoreKV+ with anchors and the adaptive amount gains up to 23.0 points under the same restore tokens.} %
\www{}{The QA gain survives on top of the restore tokens (1.1 to 5.1 points), which themselves add 2.3 to 9.2 to \ourmethod: the two act on different sides of compression, what is selected and how much a selected subset can express.} %

\textbf{Obs.\ding{188} \ourmethod wins on the categories that \www{Web questions}{questions about a document} resemble: single-document QA and summarization.}
\www{On them the best result for both models is ours (e.g., on Llama single-doc.\ QA, \ourmethod alone scores 38.1 and RestoreKV+ + \ourmethod 43.0, against at most 37.2 for any baseline; Appendix~\ref{app:lbcat}).}{Table~\ref{tab:lbcat} breaks LongBench into its official categories.
On single-document QA and summarization, which ask about the content of Wikipedia pages, papers and reports, the best result for both models is a shaded row (e.g., on Llama single-doc.\ QA, \ourmethod alone scores 38.1 and RestoreKV+ + \ourmethod 43.0, against at most 37.2 for any baseline).} %
\www{}{Inside KVgrad, \ourmethod lifts few-shot and code by 2.8 and 3.0 on Qwen3-8B and keeps multi-document QA (41.5 vs.\ 41.7).} %

\section{Related Work}\label{sec:related}

\paragraph{\textbf{Query-aware KV compression.}}
\www{H2O~\cite{h2o}, StreamingLLM~\cite{streamingllm}, SnapKV~\cite{snapkv}, PyramidKV~\cite{pyramidkv}, ChunkKV~\cite{chunkkv}, Ada-KV~\cite{adakv} and DapQ~\cite{dapq} keep heavy hitters, attention sinks, recent windows, or the entries an observation window or pseudo queries attend to.}{H2O~\cite{h2o} keeps heavy hitters of accumulated attention, StreamingLLM~\cite{streamingllm} keeps attention sinks and a recent window, SnapKV~\cite{snapkv} and PyramidKV~\cite{pyramidkv} vote with an observation window at the end of the prompt, ChunkKV~\cite{chunkkv} keeps semantic chunks rather than single tokens, Ada-KV~\cite{adakv} adapts the budget across heads, and DapQ~\cite{dapq} places pseudo queries at the positions of the first generated tokens.} \www{Run before the question, as Web caches must, they collapse (Table~\ref{tab:main}).}{All were designed to read the question or the queries seen so far; run before it, as a compress-once cache must, they collapse (Table~\ref{tab:main}).}

\paragraph{\textbf{Query-agnostic rehearsal scoring.}}
\www{KVzip~\cite{kvzip} introduced rehearsal scoring and argues that the whole context must be repeated.}{KVzip~\cite{kvzip} introduced scoring by rehearsal and argues explicitly that the whole context must be repeated.} \www{Its successors keep this target and improve the scorer or its cost: KVzip+ and KVzap~\cite{kvzap}, Fast KVzip~\cite{fastkvzip}, EchoPress~\cite{echopress} and KVgrad~\cite{kvgrad}.}{Its successors keep this target and improve the scorer or its cost: the KVzap paper~\cite{kvzap} introduces KVzip+, which normalizes by the value contribution, and KVzap, Fast KVzip~\cite{fastkvzip} and EchoPress~\cite{echopress} approximate the full-rehearsal score cheaply, and KVgrad~\cite{kvgrad} replaces attention with a gradient attribution.} All are instances of the \textbf{rehearse-everything default}, and their scorers compose with our rehearsal. Concurrent work, KV$^2$~\cite{kvsquared}, rehearses KeyDiff anchors~\cite{keydiff} at a fixed ratio $\rho$; it is the \textbf{fixed-ratio special case} of \emph{rehearse as much as there is}\www{}{, and like full-context rehearsal it rehearses a copy of the context}, and its own observation that the best $\rho$ drifts with the budget is what the principle predicts.
\www{InfiniPot~\cite{infinipot} and TaskPress~\cite{taskpress} rehearse a short fixed prompt or an LLM-written task description instead; that the latter already beats full-context rehearsal is consistent with \emph{rehearse what will be read}.}{InfiniPot~\cite{infinipot} scores the cache with a short fixed ``catalyst'' prompt, and TaskPress~\cite{taskpress} replaces the rehearsal with a two-to-three-sentence task description written by an LLM; that the TaskPress probe already beats full-context rehearsal at 50--75\% eviction is consistent with \emph{rehearse what will be read}, though both probes are fixed in advance rather than drawn from the document.}

\paragraph{\textbf{Capacity-side and learned caches.}}
\www{RestoreKV~\cite{restorekv} trains restore tokens that enlarge what a kept subset can express. Cartridges~\cite{cartridges}, Cartridges++~\cite{cartridgespp}, Attention Matching~\cite{attentionmatching}, ARC-KV~\cite{arckv}, PatchKV~\cite{patchkv} and latent context models~\cite{latentcontext} train or fit a compact cache for the same compress-once, query-many setting; Attention Matching and ARC-KV draw most reference queries from a verbatim repeat of the context; the former's self-study is closest to our QA rehearsal.}{RestoreKV~\cite{restorekv} enlarges what a kept subset can express with restore tokens trained on KVzip+ selections, rather than choosing the subset. Cartridges~\cite{cartridges} distill a document into a trained cache via self-study, Cartridges++~\cite{cartridgespp} guards it against off-context derailment, Attention Matching~\cite{attentionmatching} fits compact keys and values to reference queries, ARC-KV~\cite{arckv} amortizes the anchor search of such compaction, and PatchKV~\cite{patchkv} compensates eviction in weight space, and latent context models~\cite{latentcontext} train an encoder to compress the input itself.} \www{}{These methods share our compress-once, query-many setting; Attention Matching and ARC-KV draw most of their reference queries from a verbatim repeat of the context, another instance of the rehearse-everything default; Attention Matching adds a few self-study passes in which the model asks and answers its own questions, the closest relative of our question--answer rehearsal.} \www{}{\ourmethod keeps native KV and trains nothing.}

\ifgeneral
\paragraph{\textbf{Study strategies for language models.}}
Retrieval practice, answering one's own questions, is among the most effective ways for people to learn, and rereading among the least~\cite{roediger2006testing,dunlosky2013improving}. Language models that must learn documents into their weights benefit from the same contrast: Cartridges~\cite{cartridges} find that training on the raw corpus memorizes the text but does not generalize, while self-study conversations do, and Active Reading~\cite{activereading} lets the model generate its own study strategies, including active recall, and synthetic continued pretraining~\cite{entigraph} rewrites a small corpus into diverse synthetic text before learning it. Predicting the questions a document will be asked also has a precedent in retrieval, where doc2query~\cite{doc2query} expands documents with them. \ourmethod brings the contrast to a setting where nothing is trained: the rehearsal selects which KV entries stay, and the model quizzes itself instead of rereading.
\fi

\paragraph{\textbf{KV cache reuse \www{for Web and RAG serving}{for retrieval and serving}.}}
Prompt Cache~\cite{promptcache}, CacheBlend~\cite{cacheblend}, RAGCache~\cite{ragcache} and TurboRAG~\cite{turborag} \www{precompute and reuse document KV caches}{precompute document KV caches and reuse them across requests}, LMCache~\cite{lmcache} and Mooncake~\cite{mooncake} store them across GPU, CPU and disk tiers, \www{full-size caches whose storage grows with the corpus, and SCBench~\cite{scbench} shows that they must be compressed query-agnostically;}{solving fusion and latency for full-size caches whose storage grows with the corpus; SCBench~\cite{scbench} shows that compression conditioned on one request degrades when the same cache serves later ones, which is why such caches must be compressed query-agnostically;} CacheGen~\cite{cachegen} encodes \www{caches compactly}{stored caches into compact bitstreams}, \www{which multiplies with eviction;}{and since encoding shrinks each kept entry and eviction how many are kept, the two multiply;} \ourmethod is the eviction that lets them scale\www{}{ to large corpora}.

\paragraph{\textbf{Other importance signals.}}
\www{LORE-KV~\cite{lorekv}, Lookahead Q-Cache~\cite{lookaheadqcache} and SpecKV~\cite{speckv} score entries by the attention of a drafted response to a known request, the query-aware counterpart of \emph{rehearse what will be read}; Expected Attention~\cite{expectedattention}, CapKV~\cite{capkv}, Behavior-Preserving KV compression~\cite{behaviorpreserving}, Judge Q~\cite{judgeq}, GVote~\cite{gvote}, Compactor~\cite{compactor} and Back from the Future~\cite{backfromfuture} use other signals, and LLMLingua-2~\cite{llmlingua2} compresses the text itself. }{Expected Attention~\cite{expectedattention} and CapKV~\cite{capkv} model future queries statistically, as a Gaussian over hidden states or an information-capacity objective. Behavior-Preserving KV compression~\cite{behaviorpreserving} also questions the reconstruction target of KVzip, replacing it with a next-token KL score at available query positions, LORE-KV~\cite{lorekv}, Lookahead Q-Cache~\cite{lookaheadqcache} and SpecKV~\cite{speckv} draft the response to a known request and score entries by the attention the draft pays them; that response-side queries beat prompt-side ones is the query-aware counterpart of \emph{rehearse what will be read}. Judge Q~\cite{judgeq} trains soft query tokens whose attention imitates that of generated responses, and GVote~\cite{gvote} samples synthetic queries from hidden-state statistics. Compactor~\cite{compactor} scores entries without a rehearsal, by key leverage and non-causal attention, and calibrates the keep ratio to each context. Back from the Future~\cite{backfromfuture} evicts entries that later tokens predict well, by a counter-causal pass over the cache. LLMLingua-2~\cite{llmlingua2} compresses the text itself before prefill. }\www{\ourmethod changes only what, and how much, is rehearsed.}{\ourmethod keeps rehearsal scoring and changes what, and how much, is rehearsed.}

\section{Conclusion}\label{sec:conclusion}

A KV cache that is compressed before its questions are known must decide, from the context alone, what to remember.
The rehearsal-scoring paradigm decides by rereading: it rehearses everything and remembers nothing, because a cache keeps what it rehearses, and a whole-context rehearsal fills the budget with prose that no query reads.
Two principles turn this failure into a design rule, \textit{rehearse what will be read} and \textit{rehearse as much as there is}, and \ourmethod realizes them without training, by letting the model quiz itself on the context.
Alone, \ourmethod is the best training-free method; inside KVgrad and RestoreKV+, it lifts each above every existing method that trains as little; and its lead grows as the budget shrinks.
Since what a cache remembers depends more on what it rehearses than on how it scores, the same principles suggest how to choose the proxy queries of any compressor that stands in for unknown questions, from token eviction to learned compaction.
Holmes's attic has no elastic walls, and neither does a compressed cache.
The skilful workman does not carry in everything he comes across; he carries in the tools the work will need, and as many as the work needs.

\bibliographystyle{ACM-Reference-Format}
\bibliography{refs_A,refs_B,refs_C,refs_D}

\appendix
\raggedbottom

\section{Implementation Details}\label{app:impl}

\paragraph{\textbf{Framework.}}
All methods run in kvpress~\cite{kvpress} on the same models, prompts and splits. Context is truncated to 32K tokens on LongBench and LooGLE. %
KVgrad, KVzip+ and RestoreKV+ use their kvpress implementations; KV$^2$ is implemented in the same scoring and eviction code as KVzip+.
\www{}{Fixed-block retention in the style of ChunkKV~\cite{chunkkv}, applied to KVzip+ scores, reaches only 22.5 on Qwen3-8B RULER-4K, so we leave it out of the tables.} %

\paragraph{\textbf{Development and test splits.}}
On RULER, the development split is a random 10\% of each context length (seed 42), and the test split draws the same number of questions from the remainder (seed 7). %
On LongBench, each task is shuffled (seed 42); the first 25 questions form the development split and the next 75 (Qwen) or 50 (Llama) form the test split. %

\paragraph{\textbf{Rehearsal scoring and anchors.}}
Every rehearsal-based method shares the KVzip+ scoring and eviction of Eq.~\eqref{eq:score}, combining the scores of several rehearsed texts by an element-wise maximum. Anchors and the content-adaptive amount follow Section~\ref{sec:attic}, with chunks of 4096 tokens rehearsed and scored separately.

\paragraph{\textbf{Self-generated QA rehearsal.}}\label{app:prompt}
The QA pairs are decoded greedily from the prompt below, appended to the context after two newlines and the same for all models and benchmarks:
\begin{quote}
\small\ttfamily\raggedright
Write 6 questions a reader is likely to ask about the text above. After each question, answer it by quoting the exact words from the text that contain the answer. Format each as:\\
Q: ...\\
A: ...
\end{quote}
The header $\textsc{H}$ is ``Questions a reader may ask about the previous context, with answers quoted from it:''. %

\paragraph{\textbf{Host methods.}}
RestoreKV+ + \ourmethod adds the restore tokens and LoRA adapter that the RestoreKV authors released for Qwen3-8B and Llama-3.1-8B, unchanged, to \ourmethod's selection. They were trained on KVzip+ selections; retraining them on ours for 1000 steps does not help (83.6 vs.\ 84.3, Qwen3-8B RULER-4K development split). %

\ifgeneral\section{The \ourmethod Algorithm}\label{app:alg}

\begin{algorithm}[H]
\small
\caption{\ourmethod: query-agnostic, one-shot compression of a document's KV cache}
\label{alg:attic}
\begin{algorithmic}[1]
\Require document $C$; model $f$; keep ratio $r$; host scoring rule $\phi$ (default: Eq.~\eqref{eq:score}); optional restore tokens $\mathcal{T}$
\Ensure retained cache $\mathrm{KV}_{\mathcal{S}}$
\State Prefill $C$ to obtain the entries $\mathcal{E}$ and KeyDiff scores $\kappa$
\State $Q\gets f(C\oplus\textsc{QAPrompt})$ \Comment{6 quoted QA pairs}
\State $s\gets\mathbf{0}$
\For{each chunk $C_j$ of $C$}
  \State $\mathcal{A}_j\gets$ top-$\lfloor r|C_j|\rfloor$ positions of $C_j$ by $\kappa$ \Comment{what}
  \State $m_j\gets$ \#tokens of the sentences that contain $\mathcal{A}_j$ \Comment{how much}
  \State $R_j\gets$ top-$m_j$ positions of $C_j$ by $\kappa$, in order
  \State $s\gets\max\big(s,\ \phi(R_j;C_j),\ \phi(\textsc{H}\oplus Q;C_j)\big)$
\EndFor
\State $\mathcal{S}\gets\mathrm{Top}_r(s)$ over all layers and heads
\State \Return $\mathrm{KV}_{\mathcal{S}}$, with $\mathcal{T}$ attached if given
\end{algorithmic}
\end{algorithm}

\fi

\section{Selecting $\rho$ for KV$^2$}\label{app:rho}

KV$^2$ rehearses a fixed fraction $\rho$ of anchor tokens per chunk. Its authors set $\rho$ to the keep ratio by default; since this default is far from optimal on our development splits, we give KV$^2$ a tuned ratio. For each model and keep ratio, we evaluate a grid of $\rho$ on the RULER-4K development split and use the best value on every benchmark (RULER at 4K--64K tokens, LongBench and LooGLE), as \ourmethod uses one configuration everywhere; at the 3\% keep ratio and in the budget sweep, $\rho$ is re-selected for each keep ratio (Table~\ref{tab:app-rho}). The selected ratios are 0.35 for Qwen3-8B, 0.5 for Llama-3.1-8B and 0.5 for Qwen3-4B at a 5\% keep ratio, 0.25 for Qwen3-8B at 3\%, and 0.9 at 10\%, 20\% and 30\%. %
The rows marked $^\ddagger$ instead re-select $\rho$ on the LongBench development split, 0.15 for Llama-3.1-8B and 0.5 for Qwen3-8B, which requires knowing in advance which benchmark a document comes from. %
For Qwen3-14B, $\rho$ is chosen from $\{0.25, 0.35, 0.5\}$ in the same way, and $\rho{=}0.5$ is selected. %
\ourmethod uses no such search: its anchor ratio is the keep ratio in every setting.
\www{}{Table~\ref{tab:app-rhoscan} lists the test-split $\rho$ scan plotted in Figure~\ref{fig:rho}; it is used only for that analysis and not for selecting $\rho$.}

\begin{table}[H]
\caption{KV$^2$ development results for each candidate $\rho$ (RULER-4K development split; LB: LongBench development split, 14 natural-text tasks, used only for the $^\ddagger$ rows); the selected value is in \textbf{bold}.}
\label{tab:app-rho}
\footnotesize
\setlength{\tabcolsep}{4pt}\setlength{\aboverulesep}{0pt}\setlength{\belowrulesep}{0pt}\renewcommand{\arraystretch}{1.1}
\begin{tabular}{l p{0.68\columnwidth}}
\toprule
\headrow \textbf{Setting} & \textbf{$\boldsymbol{\rho}$: development score} \\
\midrule
Qwen3-8B, 5\% & 0.05: 68.0;\ 0.25: 74.0;\ \textbf{0.35: 74.6};\ 0.5: 73.9 \\ %
Llama-3.1-8B, 5\% & 0.05: 68.1;\ 0.1: 70.6;\ 0.25: 73.7;\ 0.35: 73.0;\ \textbf{0.5: 73.9};\ 0.7: 68.6 \\ %
Llama-3.1-8B, LB & 0.05: 37.2;\ \textbf{0.15: 39.7};\ 0.5: 37.1 \\ %
Qwen3-8B, LB & 0.05: 37.9;\ 0.15: 38.5;\ 0.35: 38.0;\ \textbf{0.5: 38.6} \\ %
Qwen3-4B, 5\% & 0.25: 70.85;\ 0.35: 70.34;\ \textbf{0.5: 70.87};\ 0.7: 68.96 \\ %
Qwen3-14B, 5\% & 0.25: 76.5;\ 0.35: 79.4;\ \textbf{0.5: 80.1};\ 0.7: 79.1 \\ %
Qwen3-8B, 3\% & 0.15: 64.71;\ \textbf{0.25: 65.33};\ 0.35: 62.41 \\ %
Qwen3-8B, 10\% & 0.1: 75.5;\ 0.35: 83.6;\ 0.5: 86.0;\ 0.7: 88.7;\ \textbf{0.9: 91.0} \\ %
Qwen3-8B, 20\% & 0.2: 86.9;\ 0.5: 91.4;\ 0.7: 92.9;\ \textbf{0.9: 93.2} \\ %
Qwen3-8B, 30\% & 0.5: 94.0;\ 0.7: 94.5;\ \textbf{0.9: 94.6} \\ %
\bottomrule
\end{tabular}
\end{table}

\ifgeneral
\begin{table}[H]
\caption{KV$^2$ $\rho$ scan on the test splits (keep ratio 5\%; LongBench: 14 natural-text tasks), the data of Figure~\ref{fig:rho}. Amount: \ourmethod's content-adaptive amount without the QA rehearsal. ``--'': not run.}
\label{tab:app-rhoscan}
\footnotesize
\setlength{\tabcolsep}{5pt}\setlength{\aboverulesep}{0pt}\setlength{\belowrulesep}{0pt}\renewcommand{\arraystretch}{1.1}
\begin{tabular}{l ccccc c >{\columncolor{oursbg}}c}
\toprule
\headrow & \multicolumn{5}{c}{\textbf{KV$^2$, $\boldsymbol{\rho =}$}} & & \\
\headrow \textbf{Setting} & \textbf{0.05} & \textbf{0.15} & \textbf{0.25} & \textbf{0.35} & \textbf{0.5} & \textbf{Amount} & \textbf{\ourmethod} \\
\midrule
Qwen3-8B R-4K & 70.3 & 72.6 & -- & 76.8 & 75.9 & 78.8 & \textbf{82.1} \\ %
Qwen3-8B LB & 37.1 & 38.7 & -- & 38.6 & 39.1 & 38.6 & \textbf{40.6} \\ %
Llama R-4K & 70.2 & 72.5 & 73.9 & -- & 74.8 & 75.6 & \textbf{77.2} \\ %
Llama LB & 38.5 & 39.1 & 38.7 & -- & 35.0 & 38.3 & \textbf{39.3} \\ %
\bottomrule
\end{tabular}
\end{table}
\fi

\ifgeneral\else\begin{table*}[h]
\caption{Per-subtask results, Qwen3-8B (test split).}\label{tab:app-ruler-www-0}
\footnotesize\setlength{\tabcolsep}{3.2pt}\setlength{\aboverulesep}{0pt}\setlength{\belowrulesep}{0pt}
\begin{tabular}{lcccccccccccccc}
\toprule
\headrow \textbf{Method} & \textbf{S1} & \textbf{S2} & \textbf{S3} & \textbf{MK1} & \textbf{MK2} & \textbf{MK3} & \textbf{MV} & \textbf{MQ} & \textbf{VT} & \textbf{CWE} & \textbf{FWE} & \textbf{QA1} & \textbf{QA2} & \textbf{Avg.} \\
\midrule
\headrow \multicolumn{15}{l}{\textit{RULER-4K}} \\
Full cache & 100.0 & 100.0 & 100.0 & 100.0 & 100.0 & 100.0 & 100.0 & 100.0 & 100.0 & 99.2 & 96.8 & 84.0 & 75.0 & 96.5 \\
KVzip+ & 100.0 & 63.8 & 61.4 & 42.0 & 60.5 & 9.4 & 42.1 & 56.3 & 100.0 & 58.1 & 68.8 & 48.0 & 51.9 & 58.6 \\
KV$^2$ & 100.0 & 95.7 & 100.0 & 94.0 & 34.9 & 1.9 & 95.8 & 96.6 & 100.0 & 85.2 & 91.4 & 58.0 & 44.2 & 76.8 \\
\oursrow \ourmethod & 100.0 & 100.0 & 100.0 & 98.0 & 51.2 & 32.1 & 97.1 & 98.9 & 100.0 & 92.9 & 93.0 & 58.0 & 46.2 & 82.1 \\
KVgrad & 100.0 & 85.1 & 95.5 & 74.0 & 41.9 & 20.8 & 90.0 & 85.2 & 99.3 & 81.3 & 74.7 & 34.0 & 53.9 & 72.0 \\
\oursrow \quad + \ourmethod & 100.0 & 100.0 & 100.0 & 100.0 & 74.4 & 54.7 & 98.3 & 98.9 & 100.0 & 79.4 & 89.8 & 46.0 & 46.2 & 83.7 \\
RestoreKV+$^\dagger$ & 100.0 & 74.5 & 77.3 & 68.0 & 93.0 & 83.0 & 72.5 & 92.6 & 100.0 & 62.5 & 74.7 & 64.0 & 53.9 & 78.2 \\
\oursrow \quad + \ourmethod & 100.0 & 100.0 & 100.0 & 100.0 & 72.1 & 81.1 & 99.2 & 98.9 & 100.0 & 95.8 & 94.1 & 62.0 & 51.9 & 88.9 \\
\midrule
\headrow \multicolumn{15}{l}{\textit{RULER-16K}} \\
Full cache & 100.0 & 100.0 & 100.0 & 98.0 & 100.0 & 100.0 & 100.0 & 100.0 & 100.0 & 84.2 & 92.5 & 72.0 & 69.2 & 93.5 \\
KVzip+ & 100.0 & 80.9 & 88.6 & 46.0 & 23.3 & 7.6 & 57.9 & 69.9 & 100.0 & 20.6 & 89.3 & 42.0 & 55.8 & 60.1 \\
KV$^2$ & 100.0 & 100.0 & 100.0 & 90.0 & 34.9 & 9.4 & 94.6 & 98.9 & 100.0 & 50.0 & 93.0 & 42.0 & 55.8 & 74.5 \\
\oursrow \ourmethod & 100.0 & 100.0 & 100.0 & 96.0 & 27.9 & 7.6 & 96.3 & 99.4 & 100.0 & 67.9 & 91.4 & 38.0 & 51.9 & 75.1 \\
KVgrad & 100.0 & 91.5 & 97.7 & 96.0 & 51.2 & 20.8 & 87.5 & 93.8 & 100.0 & 55.2 & 88.2 & 38.0 & 51.9 & 74.7 \\
\oursrow \quad + \ourmethod & 100.0 & 100.0 & 100.0 & 100.0 & 65.1 & 34.0 & 95.4 & 100.0 & 100.0 & 77.3 & 90.3 & 32.0 & 53.9 & 80.6 \\
RestoreKV+$^\dagger$ & 100.0 & 95.7 & 95.5 & 56.0 & 55.8 & 50.9 & 74.2 & 91.5 & 100.0 & 21.9 & 90.3 & 56.0 & 65.4 & 73.3 \\
\oursrow \quad + \ourmethod & 100.0 & 100.0 & 100.0 & 98.0 & 53.5 & 47.2 & 97.5 & 99.4 & 100.0 & 66.3 & 93.6 & 40.0 & 51.9 & 80.6 \\
\midrule
\headrow \multicolumn{15}{l}{\textit{RULER-4K, keep ratio 3\%}} \\
Full cache & 100.0 & 100.0 & 100.0 & 100.0 & 100.0 & 100.0 & 100.0 & 100.0 & 100.0 & 99.2 & 96.8 & 84.0 & 75.0 & 96.5 \\
KVzip+ & 100.0 & 31.9 & 25.0 & 6.0 & 7.0 & 0.0 & 23.8 & 11.4 & 100.0 & 6.3 & 19.4 & 38.0 & 40.4 & 31.5 \\
KV$^2$ & 100.0 & 83.0 & 100.0 & 82.0 & 14.0 & 0.0 & 90.0 & 88.1 & 100.0 & 59.4 & 82.3 & 38.0 & 34.6 & 67.0 \\
\oursrow \ourmethod & 100.0 & 100.0 & 100.0 & 94.0 & 14.0 & 1.9 & 92.9 & 96.0 & 100.0 & 86.7 & 87.6 & 44.0 & 36.5 & 73.4 \\
KVgrad & 100.0 & 48.9 & 54.6 & 64.0 & 18.6 & 3.8 & 66.7 & 59.1 & 97.0 & 57.5 & 53.2 & 28.0 & 44.2 & 53.5 \\
\oursrow \quad + \ourmethod & 100.0 & 100.0 & 100.0 & 94.0 & 18.6 & 0.0 & 94.6 & 97.2 & 99.6 & 54.2 & 85.5 & 38.0 & 36.5 & 70.6 \\
RestoreKV+$^\dagger$ & 100.0 & 53.2 & 25.0 & 32.0 & 41.9 & 11.3 & 55.4 & 46.0 & 100.0 & 20.4 & 53.8 & 36.0 & 44.2 & 47.6 \\
\oursrow \quad + \ourmethod & 100.0 & 100.0 & 100.0 & 98.0 & 18.6 & 17.0 & 96.3 & 98.3 & 100.0 & 85.8 & 89.8 & 42.0 & 38.5 & 75.7 \\
\bottomrule
\end{tabular}
\end{table*}

\begin{table*}[h]
\caption{Per-subtask results, Llama-3.1-8B, RULER-4K (test split).}\label{tab:app-ruler-www-1}
\footnotesize\setlength{\tabcolsep}{3.2pt}\setlength{\aboverulesep}{0pt}\setlength{\belowrulesep}{0pt}
\begin{tabular}{lcccccccccccccc}
\toprule
\headrow \textbf{Method} & \textbf{S1} & \textbf{S2} & \textbf{S3} & \textbf{MK1} & \textbf{MK2} & \textbf{MK3} & \textbf{MV} & \textbf{MQ} & \textbf{VT} & \textbf{CWE} & \textbf{FWE} & \textbf{QA1} & \textbf{QA2} & \textbf{Avg.} \\
\midrule
Full cache & 100.0 & 100.0 & 100.0 & 100.0 & 100.0 & 100.0 & 100.0 & 100.0 & 99.3 & 99.6 & 96.8 & 84.0 & 65.4 & 95.8 \\
KVzip+ & 100.0 & 93.6 & 72.7 & 76.0 & 93.0 & 11.3 & 57.5 & 83.5 & 97.0 & 8.1 & 76.3 & 40.0 & 51.9 & 66.2 \\
KV$^2$ & 100.0 & 97.9 & 100.0 & 98.0 & 76.7 & 5.7 & 80.0 & 93.2 & 98.1 & 29.8 & 89.3 & 54.0 & 50.0 & 74.8 \\
\oursrow \ourmethod & 100.0 & 100.0 & 97.7 & 98.0 & 51.2 & 9.4 & 90.4 & 98.9 & 98.5 & 68.1 & 94.6 & 50.0 & 46.2 & 77.2 \\
KVgrad & 100.0 & 95.7 & 93.2 & 98.0 & 97.7 & 54.7 & 85.8 & 94.9 & 98.9 & 29.4 & 87.1 & 24.0 & 42.3 & 77.1 \\
\oursrow \quad + \ourmethod & 100.0 & 100.0 & 100.0 & 100.0 & 60.5 & 56.6 & 99.2 & 100.0 & 98.9 & 44.2 & 88.2 & 48.0 & 48.1 & 80.3 \\
RestoreKV+$^\dagger$ & 100.0 & 100.0 & 95.5 & 94.0 & 100.0 & 92.5 & 89.2 & 97.7 & 99.6 & 18.5 & 84.4 & 78.0 & 63.5 & 85.6 \\
\oursrow \quad + \ourmethod & 100.0 & 100.0 & 97.7 & 100.0 & 65.1 & 81.1 & 97.5 & 100.0 & 99.6 & 72.7 & 93.6 & 58.0 & 57.7 & 86.4 \\
\bottomrule
\end{tabular}
\end{table*}
\begin{table*}[h]
\caption{Per-task LongBench results, Qwen3-8B (test split, keep ratio 5\%). Avg$_{14}$: the 14 natural-text tasks used in Table~\ref{tab:main}; Avg$_{16}$: all 16 tasks. KV$^2$ uses the $\rho$ selected on RULER, as in Table~\ref{tab:main}.}\label{tab:app-lb-sub-0}
\footnotesize\setlength{\tabcolsep}{3.1pt}\setlength{\aboverulesep}{0pt}\setlength{\belowrulesep}{0pt}
\begin{tabular}{lcccccccccccccccccc}
\toprule
\headrow \textbf{Method} & \textbf{NQA} & \textbf{Qsp} & \textbf{MFQA} & \textbf{HQA} & \textbf{2WQA} & \textbf{Musq} & \textbf{GovR} & \textbf{QMS} & \textbf{MNews} & \textbf{TREC} & \textbf{TQA} & \textbf{SSum} & \textbf{LCC} & \textbf{RB-P} & \textbf{PCnt} & \textbf{PRet} & \textbf{Avg$_{14}$} & \textbf{Avg$_{16}$} \\
\midrule
Full cache & 31.9 & 45.6 & 54.7 & 68.1 & 56.2 & 34.6 & 32.6 & 24.5 & 26.4 & 38.7 & 85.5 & 39.9 & 63.9 & 57.8 & 9.3 & 94.6 & 47.1 & 47.7 \\
KVzip+ & 21.3 & 17.2 & 33.0 & 25.8 & 13.7 & 9.4 & 22.3 & 20.5 & 12.7 & 24.0 & 79.9 & 35.1 & 23.4 & 62.4 & 2.7 & 32.0 & 28.6 & 27.2 \\
KV$^2$ & 28.4 & 30.1 & 45.1 & 48.1 & 36.7 & 21.9 & 27.2 & 21.6 & 20.4 & 38.7 & 85.8 & 35.9 & 40.6 & 60.2 & 5.3 & 62.7 & 38.6 & 38.0 \\
\oursrow \ourmethod & 27.7 & 35.5 & 48.5 & 56.4 & 41.9 & 19.7 & 28.7 & 21.8 & 19.6 & 42.7 & 85.8 & 35.4 & 44.1 & 60.4 & 2.7 & 45.8 & 40.6 & 38.5 \\
KVgrad & 24.9 & 29.5 & 38.6 & 55.8 & 45.8 & 23.5 & 25.5 & 21.2 & 19.3 & 50.7 & 85.3 & 36.8 & 48.3 & 58.6 & 9.3 & 94.1 & 40.3 & 41.7 \\
\oursrow \quad + \ourmethod & 24.1 & 26.9 & 45.4 & 57.2 & 43.1 & 24.2 & 27.9 & 22.2 & 22.0 & 59.3 & 84.5 & 37.3 & 56.0 & 56.7 & 2.7 & 79.3 & 41.9 & 41.8 \\
RestoreKV+$^\dagger$ & 19.2 & 26.9 & 34.8 & 27.3 & 20.1 & 10.4 & 26.0 & 22.6 & 21.5 & 40.0 & 78.7 & 36.5 & 38.5 & 61.2 & 2.7 & 39.1 & 33.1 & 31.6 \\
\oursrow \quad + \ourmethod & 30.0 & 35.5 & 49.5 & 53.3 & 41.8 & 22.2 & 28.5 & 23.6 & 23.0 & 50.7 & 86.7 & 36.2 & 53.0 & 59.5 & 4.0 & 44.7 & 42.4 & 40.1 \\
\bottomrule
\end{tabular}
\end{table*}

\begin{table*}[h]
\caption{Per-task LongBench results, Llama-3.1-8B (test split, keep ratio 5\%). Avg$_{14}$: the 14 natural-text tasks used in Table~\ref{tab:main}; Avg$_{16}$: all 16 tasks. KV$^2$ uses the $\rho$ selected on RULER, as in Table~\ref{tab:main}.}\label{tab:app-lb-sub-1}
\footnotesize\setlength{\tabcolsep}{3.1pt}\setlength{\aboverulesep}{0pt}\setlength{\belowrulesep}{0pt}
\begin{tabular}{lcccccccccccccccccc}
\toprule
\headrow \textbf{Method} & \textbf{NQA} & \textbf{Qsp} & \textbf{MFQA} & \textbf{HQA} & \textbf{2WQA} & \textbf{Musq} & \textbf{GovR} & \textbf{QMS} & \textbf{MNews} & \textbf{TREC} & \textbf{TQA} & \textbf{SSum} & \textbf{LCC} & \textbf{RB-P} & \textbf{PCnt} & \textbf{PRet} & \textbf{Avg$_{14}$} & \textbf{Avg$_{16}$} \\
\midrule
Full cache & 40.5 & 49.4 & 57.6 & 56.7 & 59.0 & 32.3 & 33.4 & 24.8 & 27.6 & 30.0 & 88.3 & 40.9 & 49.5 & 41.7 & 10.0 & 100.0 & 45.1 & 46.4 \\
KVzip+ & 33.8 & 19.4 & 33.8 & 25.2 & 26.5 & 11.3 & 17.9 & 20.3 & 13.5 & 10.0 & 86.2 & 33.2 & 27.5 & 55.5 & 0.0 & 34.0 & 29.6 & 28.0 \\
KV$^2$ & 30.9 & 32.9 & 39.9 & 39.0 & 40.7 & 15.5 & 24.3 & 21.7 & 21.8 & 16.0 & 87.6 & 36.6 & 29.7 & 53.3 & 2.0 & 34.0 & 35.0 & 32.9 \\
\oursrow \ourmethod & 33.3 & 35.3 & 45.8 & 45.5 & 49.2 & 25.4 & 27.7 & 22.4 & 23.9 & 26.0 & 86.9 & 39.0 & 36.0 & 53.7 & 2.0 & 50.0 & 39.3 & 37.6 \\
KVgrad & 32.0 & 32.4 & 35.1 & 39.4 & 38.0 & 25.3 & 25.9 & 22.5 & 22.8 & 44.0 & 85.3 & 40.9 & 40.8 & 46.8 & 8.0 & 98.0 & 37.9 & 39.8 \\
\oursrow \quad + \ourmethod & 29.6 & 39.3 & 48.1 & 47.7 & 42.4 & 20.5 & 27.4 & 23.1 & 24.7 & 38.0 & 82.9 & 40.7 & 40.2 & 45.7 & 10.0 & 82.0 & 39.3 & 40.1 \\
RestoreKV+$^\dagger$ & 30.7 & 31.2 & 49.8 & 38.9 & 46.6 & 18.5 & 27.5 & 22.9 & 25.8 & 52.0 & 89.2 & 16.5 & 29.0 & 47.8 & 0.0 & 46.0 & 37.6 & 35.8 \\
\oursrow \quad + \ourmethod & 33.9 & 43.2 & 51.9 & 49.8 & 54.6 & 26.5 & 28.7 & 23.3 & 26.8 & 42.0 & 85.9 & 36.5 & 29.1 & 50.4 & 4.0 & 58.0 & 41.6 & 40.3 \\
\bottomrule
\end{tabular}
\end{table*}
\fi

\section{Longer Contexts}\label{app:long}

The RULER release on Hugging Face stops at 16K tokens. We generate 32K and 64K contexts with the official RULER scripts in the kvpress configuration (Llama-3.1 tokenizer, base template, all 13 subtasks; 25 questions per subtask at 32K and 15 at 64K) and evaluate Qwen3-8B at 32K, within its native window, and Llama-3.1-8B at 64K. Nothing is tuned at these lengths: KV$^2$ keeps the $\rho$ selected on RULER-4K. \ourmethod remains the best compressed method at both lengths\www{: at 32K and 64K it scores 69.8 and 62.8, against 69.0 and 62.5 for KV$^2$, 50.6 and 53.2 for KVzip+, 49.8 and 52.1 for TaskPress, and 87.8 and 83.8 uncompressed.}{ (Table~\ref{tab:app-long}).} %

\ifgeneral
\begin{table}[H]
\caption{RULER at 32K (Qwen3-8B) and 64K (Llama-3.1-8B) tokens, keep ratio 5\%.}
\label{tab:app-long}
\footnotesize
\setlength{\aboverulesep}{0pt}\setlength{\belowrulesep}{0pt}
\renewcommand{\arraystretch}{1.1}
\begin{tabular}{l cc}
\toprule
\headrow \textbf{Method} & \textbf{RULER-32K} & \textbf{RULER-64K} \\
\midrule
Full cache & 87.8 & 83.8 \\ %
KVzip+ & 50.6 & 53.2 \\
TaskPress & 49.8 & 52.1 \\
KV$^2$ & \underline{69.0} & \underline{62.5} \\
\oursrow \ourmethod & \textbf{69.8} & \textbf{62.8} \\
\bottomrule
\end{tabular}
\end{table}
\fi

\section{Budget Sweep}\label{app:budget}

Table~\ref{tab:app-budget} gives the numbers behind Figure~\ref{fig:teaser} (Qwen3-8B, RULER-4K test split). The 3\% and 5\% columns are those of Table~\ref{tab:main}; KV$^2$ uses the ratio selected on the development split at each keep ratio (Table~\ref{tab:app-rho}). %

\begin{table}[H]
\caption{RULER-4K test score of Qwen3-8B across keep ratios. $^\dagger$Uses released restore tokens. Shaded rows use \ourmethod's rehearsal. \textbf{Bold}: above every baseline; \underline{underlined}: the strongest baseline.}
\label{tab:app-budget}
\footnotesize
\setlength{\tabcolsep}{5pt}\setlength{\aboverulesep}{0pt}\setlength{\belowrulesep}{0pt}\renewcommand{\arraystretch}{1.1}
\begin{tabular}{l ccccc}
\toprule
\headrow \textbf{Method} & \textbf{3\%} & \textbf{5\%} & \textbf{10\%} & \textbf{20\%} & \textbf{30\%} \\
\midrule
Full cache & 96.5 & 96.5 & 96.5 & 96.5 & 96.5 \\ %
KVzip+ & 31.5 & 58.6 & 93.0 & 96.0 & 96.2 \\ %
KV$^2$ (dev-tuned $\rho$) & \underline{67.0} & 76.8 & 92.3 & 95.9 & 95.8 \\ %
RestoreKV+$^\dagger$ & 47.6 & \underline{78.2} & \underline{94.9} & \underline{96.5} & \underline{96.4} \\ %
\midrule
\oursrow \ourmethod & \textbf{73.4} & \textbf{82.1} & 94.1 & 96.1 & 96.0 \\ %
\oursrow RestoreKV+ + \ourmethod$^\dagger$ & \textbf{75.7} & \textbf{88.9} & \textbf{95.2} & 96.4 & 96.2 \\ %
\bottomrule
\end{tabular}
\end{table}

\ifgeneral\else
\section{LongBench by Category}\label{app:lbcat}

Table~\ref{tab:lbcat} breaks the LongBench results of Table~\ref{tab:main} into LongBench's official categories.

\fi

\ifgeneral
\section{Web Document Caching}\label{app:webcache}
\begin{figure*}[t]
\centering
\includegraphics[width=\textwidth]{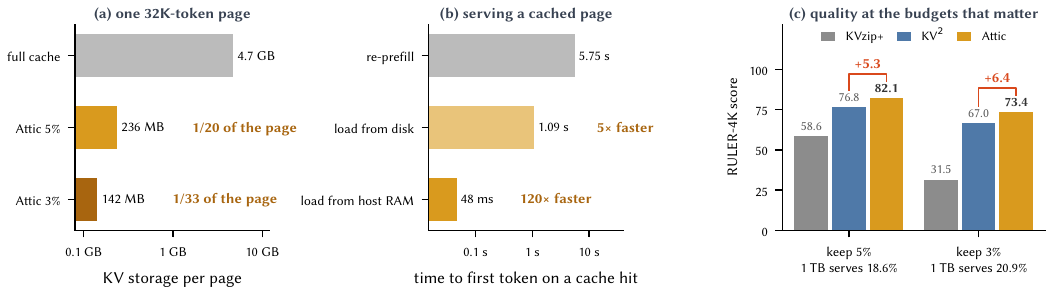}
\caption{\textbf{What a compressed Web document cache buys} (Qwen3-8B, one GPU). (a) KV storage of one 32K-token LooGLE page, stored per head. (b) Time to the first token when a request hits the cache, against prefilling the page again. (c) RULER-4K score at the two keep ratios a Web-scale cache needs; below each, the share of requests a 1\,TB cache serves under Zipf(0.8) popularity over ten million 32K-token pages.}
\label{fig:webcache}
\end{figure*}

Stored per head, as in Ada-KV~\cite{adakv}, a compressed page occupies exactly its keep ratio: over ten LooGLE pages (Figure~\ref{fig:webcache}a shows one) (20.5K tokens on average, Qwen3-8B), the KV of a page shrinks from 3.02\,GB to 151\,MB at 5\% and 91\,MB at 3\%. %
A hit skips the page's prefill: loading the compressed cache from host memory and reading the question takes 39\,ms against 3.29\,s to prefill the page again, 62--121 times faster per page, and loading it from disk is still 4.5--5.7 times faster. %
With a Zipf(0.8) popularity over ten million 32K-token pages~\cite{breslau1999web}, a 1\,TB cache of full pages serves 8.5\% of requests; at 5\% and 3\% it serves 18.6\% and 20.9\%. %
Weighted by RULER-4K accuracy, full pages answer 8.2 of 100 requests correctly from the cache; KVzip+ and KV$^2$ peak at 10\% (14.6 and 14.5), while \ourmethod keeps rising, to 15.2 at 5\% and 15.4 at 3\%, where it leads KV$^2$ by 5.3 and 6.4 points (Figure~\ref{fig:webcache}c). %
The cache is built once: with the QA pairs decoded from the prefill that compression already computes, compressing a 32K-token page costs about four prefills of it, as with KVzip+, and four hits repay it. %
We time on one GPU, alone: the full-prefill path runs the pipeline on the page and the question; the hit paths copy a cache of the compressed size from host memory or from disk (page cache dropped) and read the question. A uniform per-layer layout, padded to the head that keeps most, would take 4.5 times the per-head size. %
\fi

\section{Per-Subtask RULER Results}\label{app:ruler}

\www{Tables~\ref{tab:app-ruler-www-0} and~\ref{tab:app-ruler-www-1} break the Qwen3-8B and Llama-3.1-8B RULER settings}{Tables~\ref{tab:app-ruler-sub-0}--\ref{tab:app-ruler-sub-5} break each RULER setting} of Table~\ref{tab:main} down into \www{their}{its} 13 subtasks; the Avg.\ column is the number reported there. \www{}{KV$^2$ at its default ratio ($\rho{=}0.05$) appears in Table~\ref{tab:app-rhoscan}.} %

\ifgeneral\begin{table*}[h]
\caption{Per-subtask results, Qwen3-8B, RULER-4K (test split).}\label{tab:app-ruler-sub-0}
\footnotesize\setlength{\tabcolsep}{3.2pt}\setlength{\aboverulesep}{0pt}\setlength{\belowrulesep}{0pt}
\begin{tabular}{lcccccccccccccc}
\toprule
\headrow \textbf{Method} & \textbf{S1} & \textbf{S2} & \textbf{S3} & \textbf{MK1} & \textbf{MK2} & \textbf{MK3} & \textbf{MV} & \textbf{MQ} & \textbf{VT} & \textbf{CWE} & \textbf{FWE} & \textbf{QA1} & \textbf{QA2} & \textbf{Avg.} \\
\midrule
Full cache & 100.0 & 100.0 & 100.0 & 100.0 & 100.0 & 100.0 & 100.0 & 100.0 & 100.0 & 99.2 & 96.8 & 84.0 & 75.0 & 96.5 \\
KVzip+ & 100.0 & 63.8 & 61.4 & 42.0 & 60.5 & 9.4 & 42.1 & 56.3 & 100.0 & 58.1 & 68.8 & 48.0 & 51.9 & 58.6 \\
KV$^2$ & 100.0 & 95.7 & 100.0 & 94.0 & 34.9 & 1.9 & 95.8 & 96.6 & 100.0 & 85.2 & 91.4 & 58.0 & 44.2 & 76.8 \\
\oursrow \ourmethod & 100.0 & 100.0 & 100.0 & 98.0 & 51.2 & 32.1 & 97.1 & 98.9 & 100.0 & 92.9 & 93.0 & 58.0 & 46.2 & 82.1 \\
KVgrad & 100.0 & 85.1 & 95.5 & 74.0 & 41.9 & 20.8 & 90.0 & 85.2 & 99.3 & 81.3 & 74.7 & 34.0 & 53.9 & 72.0 \\
\oursrow \quad + \ourmethod & 100.0 & 100.0 & 100.0 & 100.0 & 74.4 & 54.7 & 98.3 & 98.9 & 100.0 & 79.4 & 89.8 & 46.0 & 46.2 & 83.7 \\
RestoreKV+$^\dagger$ & 100.0 & 74.5 & 77.3 & 68.0 & 93.0 & 83.0 & 72.5 & 92.6 & 100.0 & 62.5 & 74.7 & 64.0 & 53.9 & 78.2 \\
\oursrow \quad + \ourmethod & 100.0 & 100.0 & 100.0 & 100.0 & 72.1 & 81.1 & 99.2 & 98.9 & 100.0 & 95.8 & 94.1 & 62.0 & 51.9 & 88.9 \\
\bottomrule
\end{tabular}
\end{table*}

\begin{table*}[h]
\caption{Per-subtask results, Qwen3-8B, RULER-16K (test split).}\label{tab:app-ruler-sub-1}
\footnotesize\setlength{\tabcolsep}{3.2pt}\setlength{\aboverulesep}{0pt}\setlength{\belowrulesep}{0pt}
\begin{tabular}{lcccccccccccccc}
\toprule
\headrow \textbf{Method} & \textbf{S1} & \textbf{S2} & \textbf{S3} & \textbf{MK1} & \textbf{MK2} & \textbf{MK3} & \textbf{MV} & \textbf{MQ} & \textbf{VT} & \textbf{CWE} & \textbf{FWE} & \textbf{QA1} & \textbf{QA2} & \textbf{Avg.} \\
\midrule
Full cache & 100.0 & 100.0 & 100.0 & 98.0 & 100.0 & 100.0 & 100.0 & 100.0 & 100.0 & 84.2 & 92.5 & 72.0 & 69.2 & 93.5 \\
KVzip+ & 100.0 & 80.9 & 88.6 & 46.0 & 23.3 & 7.6 & 57.9 & 69.9 & 100.0 & 20.6 & 89.3 & 42.0 & 55.8 & 60.1 \\
KV$^2$ & 100.0 & 100.0 & 100.0 & 90.0 & 34.9 & 9.4 & 94.6 & 98.9 & 100.0 & 50.0 & 93.0 & 42.0 & 55.8 & 74.5 \\
\oursrow \ourmethod & 100.0 & 100.0 & 100.0 & 96.0 & 27.9 & 7.6 & 96.3 & 99.4 & 100.0 & 67.9 & 91.4 & 38.0 & 51.9 & 75.1 \\
KVgrad & 100.0 & 91.5 & 97.7 & 96.0 & 51.2 & 20.8 & 87.5 & 93.8 & 100.0 & 55.2 & 88.2 & 38.0 & 51.9 & 74.7 \\
\oursrow \quad + \ourmethod & 100.0 & 100.0 & 100.0 & 100.0 & 65.1 & 34.0 & 95.4 & 100.0 & 100.0 & 77.3 & 90.3 & 32.0 & 53.9 & 80.6 \\
RestoreKV+$^\dagger$ & 100.0 & 95.7 & 95.5 & 56.0 & 55.8 & 50.9 & 74.2 & 91.5 & 100.0 & 21.9 & 90.3 & 56.0 & 65.4 & 73.3 \\
\oursrow \quad + \ourmethod & 100.0 & 100.0 & 100.0 & 98.0 & 53.5 & 47.2 & 97.5 & 99.4 & 100.0 & 66.3 & 93.6 & 40.0 & 51.9 & 80.6 \\
\bottomrule
\end{tabular}
\end{table*}

\begin{table*}[h]
\caption{Per-subtask results, Qwen3-8B, RULER-4K, keep ratio 3\% (test split).}\label{tab:app-ruler-sub-2}
\footnotesize\setlength{\tabcolsep}{3.2pt}\setlength{\aboverulesep}{0pt}\setlength{\belowrulesep}{0pt}
\begin{tabular}{lcccccccccccccc}
\toprule
\headrow \textbf{Method} & \textbf{S1} & \textbf{S2} & \textbf{S3} & \textbf{MK1} & \textbf{MK2} & \textbf{MK3} & \textbf{MV} & \textbf{MQ} & \textbf{VT} & \textbf{CWE} & \textbf{FWE} & \textbf{QA1} & \textbf{QA2} & \textbf{Avg.} \\
\midrule
Full cache & 100.0 & 100.0 & 100.0 & 100.0 & 100.0 & 100.0 & 100.0 & 100.0 & 100.0 & 99.2 & 96.8 & 84.0 & 75.0 & 96.5 \\
KVzip+ & 100.0 & 31.9 & 25.0 & 6.0 & 7.0 & 0.0 & 23.8 & 11.4 & 100.0 & 6.3 & 19.4 & 38.0 & 40.4 & 31.5 \\
KV$^2$ & 100.0 & 83.0 & 100.0 & 82.0 & 14.0 & 0.0 & 90.0 & 88.1 & 100.0 & 59.4 & 82.3 & 38.0 & 34.6 & 67.0 \\
\oursrow \ourmethod & 100.0 & 100.0 & 100.0 & 94.0 & 14.0 & 1.9 & 92.9 & 96.0 & 100.0 & 86.7 & 87.6 & 44.0 & 36.5 & 73.4 \\
KVgrad & 100.0 & 48.9 & 54.6 & 64.0 & 18.6 & 3.8 & 66.7 & 59.1 & 97.0 & 57.5 & 53.2 & 28.0 & 44.2 & 53.5 \\
\oursrow \quad + \ourmethod & 100.0 & 100.0 & 100.0 & 94.0 & 18.6 & 0.0 & 94.6 & 97.2 & 99.6 & 54.2 & 85.5 & 38.0 & 36.5 & 70.6 \\
RestoreKV+$^\dagger$ & 100.0 & 53.2 & 25.0 & 32.0 & 41.9 & 11.3 & 55.4 & 46.0 & 100.0 & 20.4 & 53.8 & 36.0 & 44.2 & 47.6 \\
\oursrow \quad + \ourmethod & 100.0 & 100.0 & 100.0 & 98.0 & 18.6 & 17.0 & 96.3 & 98.3 & 100.0 & 85.8 & 89.8 & 42.0 & 38.5 & 75.7 \\
\bottomrule
\end{tabular}
\end{table*}

\begin{table*}[h]
\caption{Per-subtask results, Llama-3.1-8B, RULER-4K (test split).}\label{tab:app-ruler-sub-3}
\footnotesize\setlength{\tabcolsep}{3.2pt}\setlength{\aboverulesep}{0pt}\setlength{\belowrulesep}{0pt}
\begin{tabular}{lcccccccccccccc}
\toprule
\headrow \textbf{Method} & \textbf{S1} & \textbf{S2} & \textbf{S3} & \textbf{MK1} & \textbf{MK2} & \textbf{MK3} & \textbf{MV} & \textbf{MQ} & \textbf{VT} & \textbf{CWE} & \textbf{FWE} & \textbf{QA1} & \textbf{QA2} & \textbf{Avg.} \\
\midrule
Full cache & 100.0 & 100.0 & 100.0 & 100.0 & 100.0 & 100.0 & 100.0 & 100.0 & 99.3 & 99.6 & 96.8 & 84.0 & 65.4 & 95.8 \\
KVzip+ & 100.0 & 93.6 & 72.7 & 76.0 & 93.0 & 11.3 & 57.5 & 83.5 & 97.0 & 8.1 & 76.3 & 40.0 & 51.9 & 66.2 \\
KV$^2$ & 100.0 & 97.9 & 100.0 & 98.0 & 76.7 & 5.7 & 80.0 & 93.2 & 98.1 & 29.8 & 89.3 & 54.0 & 50.0 & 74.8 \\
\oursrow \ourmethod & 100.0 & 100.0 & 97.7 & 98.0 & 51.2 & 9.4 & 90.4 & 98.9 & 98.5 & 68.1 & 94.6 & 50.0 & 46.2 & 77.2 \\
KVgrad & 100.0 & 95.7 & 93.2 & 98.0 & 97.7 & 54.7 & 85.8 & 94.9 & 98.9 & 29.4 & 87.1 & 24.0 & 42.3 & 77.1 \\
\oursrow \quad + \ourmethod & 100.0 & 100.0 & 100.0 & 100.0 & 60.5 & 56.6 & 99.2 & 100.0 & 98.9 & 44.2 & 88.2 & 48.0 & 48.1 & 80.3 \\
RestoreKV+$^\dagger$ & 100.0 & 100.0 & 95.5 & 94.0 & 100.0 & 92.5 & 89.2 & 97.7 & 99.6 & 18.5 & 84.4 & 78.0 & 63.5 & 85.6 \\
\oursrow \quad + \ourmethod & 100.0 & 100.0 & 97.7 & 100.0 & 65.1 & 81.1 & 97.5 & 100.0 & 99.6 & 72.7 & 93.6 & 58.0 & 57.7 & 86.4 \\
\bottomrule
\end{tabular}
\end{table*}

\ifgeneral
\begin{table*}[h]
\caption{Per-subtask results, Qwen3-4B, RULER-4K (test split).}\label{tab:app-ruler-sub-4}
\footnotesize\setlength{\tabcolsep}{3.2pt}\setlength{\aboverulesep}{0pt}\setlength{\belowrulesep}{0pt}
\begin{tabular}{lcccccccccccccc}
\toprule
\headrow \textbf{Method} & \textbf{S1} & \textbf{S2} & \textbf{S3} & \textbf{MK1} & \textbf{MK2} & \textbf{MK3} & \textbf{MV} & \textbf{MQ} & \textbf{VT} & \textbf{CWE} & \textbf{FWE} & \textbf{QA1} & \textbf{QA2} & \textbf{Avg.} \\
\midrule
Full cache & 100.0 & 100.0 & 100.0 & 100.0 & 100.0 & 100.0 & 100.0 & 100.0 & 100.0 & 94.0 & 88.2 & 76.0 & 71.2 & 94.6 \\
KVzip+ & 100.0 & 72.3 & 65.9 & 42.0 & 72.1 & 5.7 & 32.1 & 51.1 & 100.0 & 36.7 & 68.8 & 46.0 & 42.3 & 56.5 \\
KV$^2$ & 100.0 & 97.9 & 100.0 & 94.0 & 58.1 & 13.2 & 84.2 & 78.4 & 100.0 & 56.0 & 78.0 & 50.0 & 53.9 & 74.1 \\
\oursrow \ourmethod & 100.0 & 100.0 & 100.0 & 100.0 & 55.8 & 7.6 & 97.1 & 84.1 & 100.0 & 74.4 & 81.2 & 48.0 & 46.2 & 76.5 \\
KVgrad & 100.0 & 87.2 & 100.0 & 78.0 & 55.8 & 34.0 & 72.9 & 71.6 & 99.6 & 75.4 & 72.6 & 32.0 & 57.7 & 72.1 \\
\oursrow \quad + \ourmethod & 100.0 & 100.0 & 100.0 & 100.0 & 83.7 & 45.3 & 95.0 & 94.3 & 100.0 & 72.7 & 79.6 & 42.0 & 46.2 & 81.4 \\
\bottomrule
\end{tabular}
\end{table*}

\begin{table*}[h]
\caption{Per-subtask results, Qwen3-14B, RULER-4K (test split).}\label{tab:app-ruler-sub-5}
\footnotesize\setlength{\tabcolsep}{3.2pt}\setlength{\aboverulesep}{0pt}\setlength{\belowrulesep}{0pt}
\begin{tabular}{lcccccccccccccc}
\toprule
\headrow \textbf{Method} & \textbf{S1} & \textbf{S2} & \textbf{S3} & \textbf{MK1} & \textbf{MK2} & \textbf{MK3} & \textbf{MV} & \textbf{MQ} & \textbf{VT} & \textbf{CWE} & \textbf{FWE} & \textbf{QA1} & \textbf{QA2} & \textbf{Avg.} \\
\midrule
Full cache & 100.0 & 100.0 & 100.0 & 100.0 & 100.0 & 100.0 & 100.0 & 100.0 & 100.0 & 99.6 & 96.2 & 92.0 & 71.2 & 96.8 \\
KVzip+ & 100.0 & 87.2 & 88.6 & 86.0 & 95.4 & 34.0 & 64.2 & 86.4 & 99.6 & 60.6 & 81.2 & 66.0 & 57.7 & 77.4 \\
KV$^2$ & 100.0 & 91.5 & 100.0 & 96.0 & 76.7 & 9.4 & 95.8 & 99.4 & 100.0 & 84.2 & 91.9 & 66.0 & 42.3 & 81.0 \\
\oursrow \ourmethod & 100.0 & 100.0 & 100.0 & 100.0 & 79.1 & 62.3 & 100.0 & 100.0 & 100.0 & 87.9 & 91.9 & 68.0 & 50.0 & 87.6 \\
KVgrad & 100.0 & 95.7 & 97.7 & 88.0 & 74.4 & 45.3 & 92.1 & 77.8 & 100.0 & 90.4 & 80.1 & 34.0 & 57.7 & 79.5 \\
\oursrow \quad + \ourmethod & 100.0 & 100.0 & 100.0 & 98.0 & 83.7 & 77.4 & 100.0 & 100.0 & 100.0 & 91.7 & 90.3 & 48.0 & 57.7 & 88.2 \\
\bottomrule
\end{tabular}
\end{table*}

\fi
\fi

\section{Per-Task LongBench Results}\label{app:longbench}

Tables~\ref{tab:app-lb-sub-0} and~\ref{tab:app-lb-sub-1} give the per-task LongBench scores with the average over the 14 natural-text tasks used in the main text (Avg$_{14}$) and over all 16 English tasks (Avg$_{16}$), which adds the two synthetic tasks (passage\_count and passage\_retrieval\_en); Appendix~\ref{app:coherent} shows why coherent rehearsal reads the content of a passage rather than its label.

\ifgeneral\fi

\section{What Coherent Rehearsal Reads}\label{app:coherent}

This appendix asks what the model reads \emph{within} a rehearsal, and why \ourmethod rehearses one coherent stream (Section~\ref{sec:attic}).

\paragraph{\textbf{Coherent rehearsal reads only what it cannot predict.}}
A rehearsal score is the attention that rehearsed tokens pay back to the context. When the rehearsal is coherent text, as \ourmethod's anchor rehearsal and self-generated QA pairs are, the model continues the stream from its own local context and looks back into the document only for tokens it cannot guess: needle values, entities, numbers. These are exactly the tokens that \www{Web questions}{questions about a document} ask for, and they are kept. Structure that is \textbf{predictable but must be read} meets the opposite fate: sequentially numbered passage labels (``Paragraph~7'') and the labels of in-context examples are produced from the stream itself, draw no attention back to the document, and are lost.

Table~\ref{tab:coherent}(a) shows this on LongBench passage retrieval, whose answer is a ``Paragraph~$N$'' label. Because every sentence of a Wikipedia passage is an askable fact, the content-adaptive amount rehearses about half of the context on this task. %
As rehearsal grows from isolated anchors (KV$^2$, 5\%) into coherent text, the kept fraction of label tokens falls from about $5\times$ the 5\% base rate to the base rate itself, and the score falls with it (Qwen3-8B: 77~$\rightarrow$~40; Llama-3.1-8B: 100~$\rightarrow$~30). %

\paragraph{\textbf{Coherence, not amount, decides which tokens survive.}}
To isolate coherence, we keep \ourmethod's rehearsed positions fixed and only break the stream: the positions of each chunk are split by a stride into interleaved sparse passes, each rehearsed separately, with the per-token score taken as the maximum over passes. The rehearsed token count is unchanged, but no pass can be continued from local context, so every rehearsed token has to be read back. Table~\ref{tab:coherent}(b) shows both sides of the trade. Labels return: passage retrieval on Qwen3-8B rises from 59 to 78.7, level with KV$^2$ at its default ratio on the same split (79). %
Needles leave: RULER drops from 81.8 to 69.8, with multi-key-2 falling from 57 to 5 and multi-value from 98 to 59. %
Under a tight budget the two kinds of tokens compete for the same slots: coherent rehearsal spends the budget on what the model cannot predict, while sparse rehearsal spreads it over everything rehearsed and crowds the needles out.

\paragraph{\textbf{A cache keeps what its rehearsal reads.}}
The finding refines our mechanism from ``a cache keeps what it rehearses'' to \emph{a cache keeps what its rehearsal reads}. \www{Web questions}{Questions about a document} target facts, entities and numbers, which is what coherent rehearsal reads.

\ifgeneral\begin{table*}[t]\else\begin{table}[H]\fi
\caption{\www{What coherent rehearsal reads (development split, keep ratio 5\%). (a) Passage retrieval, first 10 questions. (b) Coherent vs.\ sparse multi-pass rehearsal. Amt.: rehearsal amount; Label: kept fraction of label tokens; PR: passage retrieval; MK: multi-key; MV: multi-value.}{What coherent rehearsal reads (development split, keep ratio 5\%). (a) Passage retrieval, first 10 questions: as rehearsal grows into coherent text, the kept fraction of ``Paragraph~$N$'' label tokens falls to the 5\% base rate and the score follows. (b) Qwen3-8B: breaking the same rehearsal into sparse passes recovers labels but loses RULER needles. Amt.: rehearsal amount (fraction of context); Label: kept fraction of label tokens; content-adaptive: \ourmethod's amount without QA rehearsal. PR: passage retrieval; MK2/MK3: multi-key 2/3; MV: multi-value.}}
\label{tab:coherent}
\footnotesize
\setlength{\tabcolsep}{2.6pt}
\setlength{\aboverulesep}{0pt}\setlength{\belowrulesep}{0pt}
\renewcommand{\arraystretch}{1.08}
\ifgeneral\begin{minipage}[t]{0.5\textwidth}\vspace{0pt}\centering\fi
\begin{tabular}{lcccccc}
\toprule
\headrow & \multicolumn{3}{c}{\textbf{Qwen3-8B}} & \multicolumn{3}{c}{\textbf{Llama-3.1-8B}} \\
\headrow \textbf{(a) Rehearsal} & \textbf{Amt.} & \textbf{Label} & \textbf{Score} & \textbf{Amt.} & \textbf{Label} & \textbf{Score} \\
\midrule
KV$^2$ (default) & 5\% & $\sim$25\% & 77 & 5\% & $\sim$20\% & 100 \\ %
KV$^2$ ($\rho{=}0.35$) & 35\% & $\sim$7\% & 50 & 35\% & $\sim$4.5\% & 70 \\ %
Content-adaptive & 53\% & $\sim$5\% & 40 & 40\% & $\sim$4\% & 30 \\ %
\bottomrule
\end{tabular}
\ifgeneral\end{minipage}\hfill
\begin{minipage}[t]{0.46\textwidth}\vspace{0pt}\centering\else\par\vspace{4pt}\fi
\setlength{\tabcolsep}{4.9pt}
\begin{tabular}{lccccc}
\toprule
\headrow \textbf{(b) Qwen3-8B} & \textbf{PR} & \textbf{RULER} & \textbf{MK2} & \textbf{MK3} & \textbf{MV} \\
\midrule
\oursrow Coherent (\ourmethod) & 59 & 81.8 & 57 & 30 & 98 \\ %
Sparse multi-pass & 78.7 & 69.8 & 5 & 7 & 59 \\ %
\bottomrule
\end{tabular}
\ifgeneral\end{minipage}\fi
\ifgeneral\end{table*}\else\end{table}\fi

\end{document}